\documentclass[review, authoryear]{elsarticle}
\usepackage[english]{babel}
\usepackage{amsmath}
\usepackage{amsfonts}
\usepackage{amssymb}
\usepackage{amsthm}
\usepackage{graphicx}
\usepackage{bm,xcolor,multirow}
\usepackage{tikz}
\usetikzlibrary{shapes,arrows}
\usepackage{comment}
\usepackage[title]{appendix}
\usepackage{algorithm}        
\usepackage{algpseudocode}
\usepackage{enumerate}
\usepackage[round]{natbib}
\usepackage[colorlinks=true,citecolor=blue,urlcolor=blue]{hyperref}
\usepackage{siunitx}
\usepackage{caption,booktabs, makecell}
\usepackage{pdflscape}
\usepackage{threeparttable,subcaption}
\usepackage{pgfplots}
\pgfplotsset{compat = 1.16}
\usepackage{pgfgantt}
\usepackage{tikz}
\usetikzlibrary{arrows.meta, automata}
\usepackage{tikz-timing}
\usepackage{tikz-qtree}
\usepackage{ifthen}
\usepackage{footnote}
\usepackage{float} 
\usepackage{pdfsync}
\usepackage{booktabs} 
\usepackage{tabularx}  
\usepackage{caption}   
\usepackage{enumitem}
\setlist[enumerate]{noitemsep} 

\newcolumntype{Y}{>{\centering\arraybackslash}X}

\newcolumntype{L}{>{$}l<{$}} 
\newcolumntype{C}{>{$}c<{$}} 
\newcolumntype{R}{>{$}r<{$}} 

\newtheorem{example}{Example}
\newtheorem*{example*}{Example}

\usepackage{multicol}
\usepackage{etoolbox}

\makeatletter
\newcommand*{\rom}[1]{\expandafter\@slowromancap\romannumeral #1@}
\makeatother

\makeatletter

\tikzstyle{decision} = [diamond, draw, fill=blue!20,
text width=6em, text badly centered, node distance=3cm, inner sep=0pt]
\tikzstyle{block} = [rectangle, draw, fill=blue!20,
text width=8.5em, text centered, rounded corners, minimum height=4em]
\tikzstyle{line} = [draw, -latex']
\tikzstyle{cloud} = [draw, ellipse,fill=red!20, node distance=3cm,
minimum height=2em]

\definecolor{green1}{rgb}{.3, .6, 0}
\makeatother

\begin{document}
\begin{frontmatter}
\title{An approach of deep reinforcement learning for maximizing the net present value of stochastic projects}
\author[shnu]{Wei Xu}
\author[shnu]{Fan Yang}
\ead{fan_yang@shnu.edu.cn}
\author[imee]{Qinyuan Cui}
\author[nwpu]{Zhi Chen\corref{cor}}
\ead{chenzhi@nwpu.edu.cn}

\address[shnu]{School of Finance and Business, Shanghai Normal University, China}
\address[imee]{School of Information and Mechanical and Electrical Engineering, Shanghai Normal University, China}
\address[nwpu]{School of Management, Northwestern Polytechnical University, Xi’an, China}

\cortext[cor]{Corresponding author. School of Management, Northwestern Polytechnical University, Xi’an, China.}

\begin{abstract}
We investigate a project with stochastic activity durations and cash flows under discrete scenarios, where activities must satisfy precedence constraints generating cash inflows and outflows. The objective is to maximize expected net present value (NPV) by accelerating inflows and deferring outflows. We formulate the problem as a discrete-time Markov Decision Process (MDP) and propose a Double Deep Q-Network (DDQN) approach.
Comparative experiments demonstrate that DDQN outperforms traditional rigid and dynamic strategies, particularly in large-scale or highly uncertain environments, exhibiting superior computational capability, policy reliability, and adaptability. 
Ablation studies further reveal that the dual-network architecture mitigates overestimation of action values, while the target network substantially improves training convergence and robustness. These results indicate that DDQN not only achieves higher expected NPV in complex project optimization but also provides a reliable framework for stable and effective policy implementation.

\end{abstract}

\begin{keyword}
project scheduling, net present value, deep reinforcement learning, stochastic project
\end{keyword}
\end{frontmatter}

%
%

\section{Introduction}
\label{sec:introduction}
Project scheduling involves systematic planning and sequencing of tasks, during project execution, to ensure that the project is completed efficiently within the shortest possible time frame. However, in today's competitive business environment, project scheduling not only depends on effective time management but must also prioritize maximizing economic benefits. The Net Present Value (NPV) is a critical measure of project value, reflecting the overall benefit after accounting for the time value of money. The NPV of a project is calculated by applying discount rates to all cash flows generated throughout the project's life cycle. Consequently, optimizing project scheduling decisions to maximize NPV has become a central objective in project management.

Projects can be categorized into two types: deterministic and stochastic. Deterministic projects refer to those in which all relevant parameters are known and fixed at the outset, whereas stochastic projects involve certain uncertain parameters (such as task durations and costs) that undergo random changes over time. This uncertainty makes project scheduling problems more complex and challenging. \cite{herroelen1997} conduct extensive research in deterministic settings, focusing on maximizing project net present value under various constraints and planning considerations. However, in real-world scenarios, due to the ambiguity of activity durations and the fluctuations in cash flows and discount rates, uncertainty is prevalent in project scheduling problems. Therefore, studying the uncertainty in project scheduling is of significant importance for improving project management efficiency and optimizing resource allocation.

Traditionally, uncertainty in project scheduling problems has been modeled as stochastic problems. \cite{wiesemann2015} further emphasize the importance of stochastic project scheduling. In stochastic project scheduling, activity durations and/or cash flows are treated as random variables, which directly result in the project net present value also being a random variable. Research on stochastic project scheduling problems not only better addresses the complexity and uncertainty of real-world projects but also provides more reliable support for project decision-making. The importance of stochastic project scheduling lies in its ability to help project managers develop more flexible and robust scheduling plans when faced with dynamic environments.

To address the challenge of maximizing the expected NPV in stochastic projects, scholars have typically employed branch-and-bound algorithms (\cite{wiesemann2010}), stochastic dynamic programming methods (\cite{creemers2015}), and heuristic algorithms, including tabu search techniques (\cite{waligra2008,zheng2018}).
However, these approaches generally face limitations such as high computational complexity, stringent requirements for accurately characterizing stochasticity, and insufficient solution stability. In particular, as problem size and the number of scenarios increase, the models are prone to the curse of dimensionality or a dramatic surge in computational burden, thereby restricting their practical applicability to large-scale projects.

Therefore, we introduces a deep reinforcement learning (DRL) approach by formulating the project scheduling problem as a discrete-time Markov Decision Process (MDP), systematically accounting for activity dependencies and the long-term impact of future decisions on the NPV, while ignoring resource constraints.  
By integrating the Double Deep Q-Network (DDQN) algorithm, it eliminates the reliance of traditional algorithms on predefined priority rules and leverages the powerful representational capacity of deep neural networks for high-dimensional state spaces to directly optimize long-term cumulative returns with NPV as the objective.
Through an end-to-end autonomous learning mechanism, the approach is capable of generating near-globally optimal scheduling policies directly in complex stochastic project environments. 
To validate the effectiveness of the proposed method, we conduct extensive simulation experiments across projects of varying sizes and uncertainty levels to systematically evaluate policy performance. The results indicate that, compared with traditional rigid and dynamic scheduling strategies, our approach exhibits markedly superior performance when dealing with large-scale and highly stochastic projects.

The remainder of this paper is organized as follows. Section~\ref{sec:realtedWork} introduces some related work on the considered problem of this paper.
Section~\ref{sec:problemStatement} provides a formal problem definition. Section~\ref{sec:solutionMethod} proposes algorithms that combine Markov decision processes with DDQN models to solve the above models. Section~\ref{sec:computationalResults} conducts extensive computational experiments and we conclude in Section~\ref{sec:conclusions}.

%
%

\section{Related work}
\label{sec:realtedWork}
This section provides an overview of the key literature relevant to the issue addressed in this paper. Section~\ref{sec:research on maximizing NPV of project scheduling} reviews the literature on the maximization of the NPV in project scheduling, and Section~\ref{sec:research on reinforcement learning in project scheduling} discusses the application of DRL in project scheduling.

\subsection{Research on maximizing NPV of project scheduling}
\label{sec:research on maximizing NPV of project scheduling}
The traditional project scheduling problem mainly focuses on the shortest duration as the optimization objective, but along with higher interest rates and expensive financing costs, maximizing the NPV is more reflective of the characteristics and value of the project.
\cite{russell1970} introduce NPV into the project scheduling problem for the first time. He developed an approximate iterative algorithm based on the Activity-on-Arc (AOA) network to maximize NPV under the constraint of no resource availability.
After that, \cite{grinold1972} states that the local optimum is the global optimum, linearises the project scheduling problem and solves it using the network simplex method. \cite{elmaghraby1990} extend  the algorithm of \cite{grinold1972}  by designing a graphical method and testing it on a large scale.

In the past few decades, most scholars focuse more on studying deterministic project scheduling problems (\cite{herroelen1997}, \cite{leyman2016}, \cite{leyman2017} and \cite{asadujjaman2021}).
In reality, the activity duration and/or cash flows of projects often exhibit uncertainty. Scholars have conducted extensive research on this issue, primarily focusing on two types of assumptions: activity durations following specific distributions and those following general distributions. 
For activity durations that are independent and exponentially distributed, \cite{buss1997} and \cite{sobel2009} extend the Russell's model with uncertainty using a continuous-time MDP. However, their proposed method only applies to small projects because the number of states in a MDP usually grows exponentially with project sizes. 
\cite{wiesemann2015} subdivide the project evolution into multiple decision periods such that the stochastic project NPV maximization problem is formulated as a discrete-time MDP with a discount rate that depends on the state and action. 
\cite{hermans2018} investigate project scheduling with NPV as the objective by utilizing a continuous-time Markov decision chain based on a rational division of the state space. It is shown that the optimal preemptive solution solves the non-preemptive case in a Markov PERT network and the performance of the algorithm is significantly improved. 
\cite{creemers2018} uses a new continuous-time Markov chain and a backward stochastic dynamic procedure to determine the optimal policy to maximize the expected net present value of the project. Experiments show that the method significantly reduces memory requirements, making it 600 times more computationally efficient.

For activity durations with general distributions, \cite{wiesemann2010} design a branch-and-bound algorithm based on the method of \cite{benati2006}, which solves the problem of maximizing the expected net present value of projects where activity durations and cash flows are described by correlated occurrence probabilities.
\cite{mohaghar2016} investigate the problem of maximizing the project's prefetched NPV by eliminating the safety floats increasing the possibility of activity duration and studied project scheduling with expected NPV maximization. 
\cite{liang2018} propose a composite robust scheduling model and developed a two-stage algorithm integrating simulated annealing and forbidden search for solving the project NPV maximization problem. 
\cite{zheng2018} construct two active scheduling time buffer optimization models and two passive scheduling models to study the project maximization NPV problem with stochastic activity duration. 
\cite{rezaei2020} develop a new scenario-based average conditional value-at-risk (CVaR) and mixed-integer planning model. This model simultaneously minimizes the NPV risk and maximizes the expected NPV, taking into account the uncertainty of activity durations and cash flows.  
\cite{peymankar2021} propose two integer linear programming formulations and develops a two-stage stochastic planning approach to construct the Benders decomposition algorithm to efficiently solve the project's expected NPV maximization problem. 
\cite{phuntsho2024} mixe approximate dynamic programming with three different metaheuristic algorithms to solve the project's maximum NPV by using their combined advantages.
These studies focus on the uncertainties in project scheduling and achieve a balance between complexity and efficiency through various models and optimization methods.

\subsection{The application of reinforcement learning in project scheduling}
\label{sec:research on reinforcement learning in project scheduling}
In recent years, unprecedented advancements in big data and artificial intelligence, particularly the innovative application of DRL in project optimization, substantially accelerate the digital transformation across various industries. Compared with conventional approaches, DRL algorithms, owing to their distinctive adaptive learning mechanisms, markedly reduce the complexity associated with manual modeling (\cite{chen2022},\cite{zhang2020} and \cite{luo2020}). Within dynamic project environments, DRL efficiently addresses high-dimensional, nonlinear optimization objectives, including net present value, and continuously enhances decision-making precision through data-driven learning.

\cite{avalos2023} applies DRL to mining development projects to optimize extraction decisions under spatial distribution uncertainty. Controlled simulation experiments based on multivariate drilling datasets verify that DRL maximizes project net present value.
\cite{yao2024} proposes a DRL model with an effective action-sampling mechanism to optimize large-scale construction projects. The model integrates graph convolutional networks for feature extraction and employs a reward shaping mechanism to accelerate convergence, outperforming traditional methods in both scheduling and rescheduling scenarios by reducing project duration and operational time.
\cite{wang2024} utilizes a DRL algorithm based on Double Deep Q-Learning with prioritized replay to select adaptive strategies for resolving global resource conflicts, addressing dynamic, decentralized, resource-constrained multi-project scheduling problems with product transfers, thereby minimizing the completion time of individual projects. Extensive simulation experiments based on real-world enterprise cases validate the applicability and superiority of the proposed method.
\cite{cai2024} combines reinforcement learning with graph neural networks (GNNs), formulates the scheduling process as a sequential decision-making problem, extracts features from the problem structure, and maps them to action probability distributions via a policy network. Proximal Policy Optimization (PPO) trains the model end-to-end; however, due to the inherent sparsity of graph-structured data, the model exhibits limited generalization capability.

%
%

\section{Problem statement}
\label{sec:problemStatement}
This paper addresses the project scheduling to maximize the expected NPV under activity logical constraints. The project is represented as a single-digraph network $G=(V,E)$, where $V$ denotes the set of project activities with nodes $V={0,1,...,n+1}$. Nodes 0 and $n+1$ are virtual activities that signify the project's start and completion, respectively, while the others are non-virtual activities. $E$ represents the set of precedence relationships between activities. A pair $(i,j) \in E$ implies that activity $j$ can commence only after the completion of activity $i$. The duration $\widetilde{d_j}$ of activity $j$ is governed by a specific discrete probability distribution. Each non-virtual activity $j$ generates a cash flow $\widetilde{c_j}$ consisting of a fixed cost $c_j^F$ and a stochastic variable cost $\widetilde{d_j} \cdot c_j^V + g_j$, where $c_j^F$ represents the fixed cost, $c_j^V$ denotes the variable cost per unit time, and $g_j$ signifies the revenue obtained upon completion of activity $j$. Generally, positive cash inflows accrue from the project upon completion of specified activities $V' \subseteq V$. Specifically, when activity $n+1 \in V'$, positive cash flow occurs upon project completion.
Since the activity cash flow $\tilde{c}_j$ depends on the random duration $\tilde{d}_j$, it is also treated as a random variable. The cash flow of some activities is positive, while that of others is negative. The project deadline is denoted by $\delta$.

The project scheduling problem with stochastic activity durations and cash flows can be formulated as a dynamic decision-making process, where the solution corresponds to a policy $\pi$ that specifies the action to be taken at each decision time. We consider project decisions to be made at discrete time points, each representing a decision instant. An action may involve initiating a set of "feasible" activities, meaning that a feasible decision is gradually constructed over time.
Apart from the problem input data, the decision maker can only utilize information available up to the current time (e.g., realized durations of completed activities) and cannot access future random outcomes in advance. This restriction is commonly referred to as the non-anticipativity constraint, which ensures that future uncertainty cannot be exploited at the current decision time.

We assume that (i) the activity durations $\tilde{d}_j$ are mutually independent; (ii) the cash flows $\tilde{c}_j$ are conditionally independent given the activity durations; (iii) cash flows are realized at the start of each activity $j$; and (iv) the policy $\pi$ is non-anticipative.
A policy $\pi$ maps a scenario $\sigma=(d^\sigma, c^\sigma)$ to the starting times $t_j^\sigma$ of all activities, where $d^\sigma=(d_j^\sigma)_{j\in V}$ and $c^\sigma=(c_j^\sigma)_{j\in V}$ represent the realized values of the random variables $\tilde{d}_j$ and $\tilde{c}_j$ under scenario $\sigma$. Let $\Sigma$ denote the set of all possible scenarios, where the probability of occurrence of each scenario $\sigma \in \Sigma$ is denoted by $p_{\sigma}$. It is assumed that all scenarios are equally likely, i.e., $p_{\sigma} = 1 / |\Sigma|$.

The mathematical formulation of the stochastic project scheduling problem for maximizing the NPV is presented as follows,
\begin{subequations}
	\begin{align}
		\max \quad & \sum_{\sigma\in \Sigma} p_\sigma \sum_{j\in V} c_j^\sigma \bullet \beta^{t_j^\sigma} 
		\label{eq:max_obj} \\[1ex]
		\text{s.t.} \quad & t_i^\sigma + d_i^\sigma \le t_j^\sigma, \quad \forall (i,j) \in E, \forall \sigma \in \Sigma 
		\label{eq:precedence} \\[1ex]
		& t_{n+1}^\sigma \le \delta, \quad \forall \sigma \in \Sigma 
		\label{eq:deadline} \\[0.5ex]
		& t_0^\sigma = 0, \quad \forall \sigma \in \Sigma 
		\label{eq:start_time} \\[0.5ex]
		& t_j^\sigma \ge 0, \quad t_j^\sigma \in \mathbb{Z}, \quad \forall j \in V, \forall \sigma \in \Sigma
		\label{eq:nonnegativity}
	\end{align}
\end{subequations}

The objective function \eqref{eq:max_obj} maximizes the project NPV by discounting all cash flows to the activity starting times based on the discount factor $\beta$.
Constraint \eqref{eq:precedence} enforces precedence feasibility among activities.
Constraint \eqref{eq:deadline} ensures that the total project duration does not exceed the maximum allowed deadline $\delta$.
Constraint \eqref{eq:start_time} specifies that the project starting time is zero.
Constraint \eqref{eq:nonnegativity} requires that all activity start times be non-negative integers.

\section{Solution method}
\label{sec:solutionMethod}
This section introduces MDP model and the DDQN algorithm. 
Subsection~\ref{sec:Markov Decision Process(MDP)} delineates the construction of the MDP, while Subsection~\ref{sec:Double Deep Q-Learning (DDQN) algorithm} discusses how the DDQN algorithm selects optimal decisions within feasible options, ultimately achieving the maximization of the expected net present value.

\subsection{Markov Decision Process(MDP)}
\label{sec:Markov Decision Process(MDP)}
An MDP is characterized by the tuple $(\mathcal{S},\mathcal{A}, p, r, \beta)$, where $\mathcal{S}$ represents the finite state space containing all possible states $s \in \mathcal{S}$, and $\mathcal{A}$ denotes the finite action space containing all possible actions $a \in \mathcal{A}$.
The transition function $p(s_{k+1}\mid s_k, a_k)$ specifies the probability of reaching state $s_{k+1}$ from state $s_k$ after taking action $a_k$, while the reward function $r(s_k, a_k)$ indicates the immediate reward obtained when action $a_k$ is executed in state $s_k$.
The discount factor $\beta \in [0,1)$ determines the present value of future rewards.
The solution to an MDP is a policy $\pi$ that specifies the action to take in each state. Among all possible policies $\Pi$, the objective is to identify the optimal policy $\pi^*$ that maximizes the expected cumulative reward.

We first formulate the MDP for the project scheduling problem. The system state at time step $k$ is defined as
\begin{equation}
	\label{eq4.1}
    s_k = (t_k,A_k,U_k,F_k,\boldsymbol{\varphi}_{k}, \boldsymbol{\psi}_{k}, \boldsymbol{x}_{k}),
\end{equation}
where $k$ represents the current time step or decision stage, and $s_k$ describes the system state at the $k$-th stage. Eq.~(\ref{eq4.1}) shows that the state vector $s_k$ consists of the following components:
(i) $t_k$, representing the current decision time, which is a discrete variable;
(ii) $A_k$, $U_k$, and $F_k$, which denote the sets of activities currently being executed, activities that have not started, and activities that have been completed, respectively; 
(iii) the activity duration vector $\boldsymbol{\varphi}_{k}$, where for each activity $j$, if it is completed, $\boldsymbol{\varphi}_{k}$ records its actual duration; if it is in progress, $\boldsymbol{\varphi}_{k}$ represents the time already spent on it; and if it has not started, $\boldsymbol{\varphi}_{k} = 0$;
(iv) the activity start time vector $\boldsymbol{\psi}_{k}$, initially set to $-1$, which becomes the actual start time once activity $j$ begins; and
(v) the current activity state vector $\boldsymbol{x}_{k}$, included to comprehensively capture the execution status of each activity and preserve information relevant to future decisions. Specifically, the encoding of $\boldsymbol{x}_{k}$ is as follows: 0 - not started, 1 - in progress, 2 - completed.

Secondly, at each decision time $t_k$, the system makes a decision by selecting an action $a_k \in \mathcal{A}(s_k)$, where $\mathcal{A}(s_k)$ denotes the set of feasible actions under state $s_k$. In project scheduling, the action $a_k$ corresponds to the decision of either starting a specific activity $j$ or initiating no activity. 
Choosing a particular action $a_k$ influences the transition probability from the current state $s_k$ to the next state $s_{k+1}$, denoted as $p(s_{k+1} \mid s_k, a_k)$.
The transition probability is Markovian, depending only on the previous state $s_k$ and action $a_k$, and not on the history of past states or actions.   
For each action in the feasible set $\mathcal{A}(s_k)$, the current policy $\pi$ determines the probability of selecting $a_k$ under state $s_k$, expressed as $\pi(a_k \mid s_k)$. It is important to note that an activity $j$ can be started only if all its immediate predecessors have been completed by the current decision time $t_k$; otherwise, the system may choose not to start any activity. Moreover, when the set of ongoing activities $A_k$ is non-empty, choosing no activity is a feasible option; otherwise, if no activities are currently in progress and no new activity is started, the project execution would be interrupted.
Considering all feasible actions, the overall probability of transitioning from state $s_k$ to $s_{k+1}$ is obtained as a weighted sum over all actions:
\begin{equation}
	\label{eq4.2}
	p(s_{k+1}\mid s_k) = \sum_{a_k \in \mathcal{A}(s_k)} \pi(a_k \mid s_k)\, p(s_{k+1}\mid s_k, a_k),
\end{equation}

Consequently, the transition equations between states depend on the chosen actions. The next system state at time step $k+1$ can be represented as:
\begin{equation}
	\label{eq4.3}
	s_{k+1} = (t_{k+1}, A_{k+1}, U_{k+1}, F_{k+1}, \boldsymbol{\varphi}_{k+1}, \boldsymbol{\psi}_{k+1}, \boldsymbol{x}_{k+1}).
\end{equation}

If the decision is to initiate a particular activity $j$, the next state is updated according to:
\begin{equation}
	\label{eq4.4}
	\begin{cases} 
		t_{k+1} = t_k, \\
		A_{k+1} = A_k \cup \{ j \}, \\
		U_{k+1} = U_k \setminus \{ j \}, \\
		F_{k+1} = F_k, \\
		\boldsymbol{\varphi}_{k+1} = \boldsymbol{\varphi}_{k}, \\
		\boldsymbol{\psi}_{k+1}(j) = t_k, \\
		\boldsymbol{x}_{k+1}(j) = 1,
	\end{cases}
\end{equation}
Eq.~(\ref{eq4.4}) describes the state-update process when a new activity is initiated. At decision time $t_{k+1} = t_k$, activity $j$ is added to the active set $A_{k+1}$ and simultaneously removed from the set of uninitiated activities $U_{k+1}$. The completed set $F_{k+1}$ and the activity duration vector $\boldsymbol{\varphi}_{k+1}$ remain unchanged from the previous state $s_k$. Additionally, in the activity start-time vector $\boldsymbol{\psi}_{k+1}$, the start time of activity $j$ is updated to $t_k$. Finally, the current activity state vector $\boldsymbol{x}_{k+1}$ is updated to reflect that activity $j$ is now in progress (encoded as 1).

When the decision is made to initiate no new activities, the system state at the next time step is updated according to the set of activities completed during the current period. Let $C_{k} \subseteq A_{k}$ denote the set of activities completed within the time interval $[t_{k}, t_{k+1})$. If $C_k = \emptyset$, no activities are completed within the current time step, and the system only advances in time. If $|C_k| = 1$, exactly one activity is completed, whereas $|C_k| > 1$ indicates that multiple activities are completed simultaneously. The state update rules are given as follows:
\begin{equation}
	\label{eq4.5}
	\begin{cases}
		t_{k+1} = t_k + 1, \\
		A_{k+1} = A_k \setminus C_k, \\
		U_{k+1} = U_k, \\
		F_{k+1} = F_k \cup C_k, \\
		\boldsymbol{\varphi}_{k+1}(j) = \boldsymbol{\varphi}_{k}(j) + 1, \quad \forall a_j \in A_k\\
		\boldsymbol{\psi}_{k+1} = \boldsymbol{\psi}_{k}, \\
		\boldsymbol{x}_{k+1}(j) = 2, \quad \forall j \in C_k.
	\end{cases}
\end{equation}

As shown in the Eq.~(\ref{eq4.5}), the system makes one decision at each discrete time step (i.e., $t_{k+1} = t_k + 1$). The set of ongoing activities is updated by removing the completed ones, i.e., $A_{k+1} = A_k \setminus C_k$. Since no new activities are initiated, the set of unstarted activities remains unchanged ($U_{k+1} = U_k$). The set of finished activities $F_{k+1}$ is updated by merging the newly completed activities $C_k$. 
Moreover, the activity duration vector $\boldsymbol{\varphi}_{k+1}$ increases by one for all executing activities, reflecting the natural progression of time, while the activity start-time vector $\boldsymbol{\psi}_{k+1}$ remains constant. Finally, for each activity $j \in C_k$, the corresponding state variable is set to $x_{k+1}(j) = 2$, indicating that the activity has been completed.

We subdivide the evolution of the project into multiple decision time steps, framing the problem of maximizing the expected NPV of stochastic project schedule as a discrete-time MDP. In this formulation, the discount factor dependent on the state and chosen actions.
At each decision time $t_k$, executing action $a_k$ in state $s_k$, which corresponds to starting activity $j$, yields an instantaneous reward $r(s_k, a_k)$. After taking this action, the system transitions randomly to the next state $s_{k+1}$ according to the transition probability $p(s_{k+1}|s_k, a_k)$. 
The instantaneous reward $r(s_k, a_k)$ represents the single-step cost, defined as the increment of NPV between the decision times $t_k$ and $t_{k+1}$.It includes the discounted fixed costs incurred by starting new activities at $t_k$, as well as the discounted variable costs accrued from the execution of ongoing activities during the interval $[t_k, t_{k+1}]$.
In the context of project scheduling, the cash flow required for each activity is considered its activity cost, and it is assumed that these cash flows are generated at the commencement of the activity. 
Accordingly, the instantaneous reward $r(s_k, a_k)$ for starting activity $j$ can be expressed as:
\begin{equation}
	\label{eq4.6}
	r(s_k, a_k) = npv_j = \beta^{\widetilde{t}_j} \cdot \widetilde{c}_j,
\end{equation}
where $\beta$ is the cash flow discount factor, $\widetilde{t}_j$ is the start time of activity $j$, and $\widetilde{c}_j$ is the corresponding cash flow. This formulation captures the present value of the cost incurred by initiating the activity at its start time.

As mentioned above, our goal is to find an optimal policy $\pi^* \in \Pi$ that maximizes the expected cumulative return, i.e., the expected NPV of project. Each policy $\pi \in \Pi$ defines a mapping from states $s_k$ to actions $a_k$. At a discrete decision time $t_k$, the state-value function under policy $\pi$ is defined as the expected cumulative return starting from state $s_k$ while following $\pi$:
\begin{equation}
	\label{eq-value}
	V^\pi(s_k) = \mathbb{E}\Bigg[ \sum_{t=k}^{\infty} \beta^{t-k} r(s_t, a_t) \,\Big|\, s_k \Bigg].
\end{equation}

Correspondingly, the state-action value function $Q^\pi(s_k, a_k)$ represents the expected cumulative return obtained by taking action $a_k$ in state $s_k$ and thereafter following policy $\pi$:
\begin{equation}
	\label{eq-action}
	Q^\pi(s_k, a_k) = \mathbb{E}\Big[ r(s_k, a_k) + \sum_{t=k+1}^{\infty} \beta^{t-k} r(s_t, a_t) \,\big|\, s_k, a_k \Big].
\end{equation}

By definition, the following relationship exists between Eq~\eqref{eq-value} and Eq~\eqref{eq-action}:
\begin{equation}
	\label{eq-v-a-relationship}
	V^\pi(s_k) = \sum_{a_k \in \mathcal{A}(s_k)} \pi(a_k \mid s_k) Q^\pi(s_k, a_k).
\end{equation}

According to the Bellman expectation equation, $Q^\pi(s_k, a_k)$ can be further written as:
\begin{equation}
	\label{eq4.14}
	Q^\pi(s_k, a_k) = r(s_k, a_k) + \beta \sum_{s_{k+1} \in \mathcal{S}} p(s_{k+1} \mid s_k, a_k) V^\pi(s_{k+1}).
\end{equation}

By combining Eq.~\eqref{eq-v-a-relationship} and \eqref{eq4.14}, we can derive the expanded forms of the Bellman expectation equations:
\begin{equation}
	\label{eqb-v}
	V^\pi(s_k) = \sum_{a_k \in \mathcal{A}(s_k)} \pi(a_k \mid s_k) \Big[ r(s_k, a_k) + \beta \sum_{s_{k+1} \in \mathcal{S}} p(s_{k+1} \mid s_k, a_k) V^\pi(s_{k+1}) \Big],
\end{equation}
\begin{equation}
	\label{eqb-q}
	Q^\pi(s_k, a_k) = r(s_k, a_k) + \beta \sum_{s_{k+1} \in \mathcal{S}} p(s_{k+1} \mid s_k, a_k) \sum_{a_{k+1} \in \mathcal{A}(s_{k+1})} \pi(a_{k+1} \mid s_{k+1}) Q^\pi(s_{k+1}, a_{k+1}).
\end{equation}

Our objective is to find an optimal policy $\pi^* \in \Pi$ such that its value function is maximal among all policies:
\begin{equation}
	\label{eq4.17}
	V^*(s_k) = V^{\pi^*}(s_k) = \sup_{\pi \in \Pi} V^\pi(s_k), \quad \forall s_k \in \mathcal{S},
\end{equation}
\begin{equation}
	\label{eq4.18}
	Q^*(s_k, a_k) = Q^{\pi^*}(s_k, a_k) = \sup_{\pi \in \Pi} Q^\pi(s_k, a_k), \quad \forall s_k \in \mathcal{S}, \, a_k \in \mathcal{A}(s_k).
\end{equation}

It is well-known that under mild assumptions, the optimal value functions exist and are the unique solutions to Eq.~\eqref{eqb-v} and \eqref{eqb-q}. The optimal policy $\pi^*$ can be expressed as:
\begin{equation}
	\label{best-policy}
	\pi^*(a_k \mid s_k) =
	\begin{cases}
		1, & \text{if } a_k \in \arg\max_{a_k \in \mathcal{A}(s_k)} Q^*(s_k, a_k), \\
		0, & \text{otherwise}.
	\end{cases}
\end{equation}

\subsection{Double Deep Q-Network (DDQN) algorithm}
\label{sec:Double Deep Q-Learning (DDQN) algorithm}
The primary approaches for solving the MDP model can be broadly categorized into two classes: dynamic programming and reinforcement learning (RL). Dynamic programming methods rely on known model information to derive the optimal policy via algorithms such as value iteration or policy iteration. In contrast, RL does not necessitate complete model knowledge, but instead learns the optimal policy through interaction with the environment.

The Q-learning algorithm is one of the classic algorithms in RL, , whose core idea is to find the optimal policy through learning a value function.
The algorithm iteratively collects experience by interacting with the environment and updates the state-action value function accordingly. Typically, the state-action value function is represented as a Q-table of dimension $|\mathcal{S}| \times |\mathcal{A}|$, where $|\mathcal{S}|$ denotes the number of states and $|\mathcal{A}|$ denotes the number of feasible actions. Specifically, the Q-table maps each state-action pair $(s, a)$, with $s \in \mathcal{S}$ and $a \in \mathcal{A}$, to a Q-value, which represents the expected cumulative reward of taking action $a$ in state $s$. The Q-values guide the agent to select optimal actions and gradually approximate the optimal policy. The update rule is given by:
\begin{equation}
	\label{eq4.16}
	Q(s_k, a_k) \gets Q(s_k, a_k) + \alpha \Big[ r_k + \beta \max_{a \in \mathcal{A}(s_{k+1})} Q(s_{k+1}, a) - Q(s_k, a_k) \Big],
\end{equation}
where $s_k$ is the current state, $a_k$ is the chosen action, $s_{k+1}$ is the next state after executing $a_k$, and $r_k$ is the immediate reward. The parameter $\alpha \in (0,1]$ is the learning rate controlling the update step size, while $\beta \in (0,1]$ is the discount factor balancing the importance of future rewards. The term $\max_{a \in \mathcal{A}} Q(s_{k+1}, a)$ represents the optimal expected value at the next state.

However, in high-dimensional state spaces, storing and updating the value function often requires substantial computational resources. Traditional Q-learning methods rely on discrete Q-tables, whose dimensionality grows exponentially with the size of the state and action spaces, making them impractical for high-dimensional problems. 
To overcome this limitation, \cite{mnih2015} proposed a value-based deep reinforcement learning algorithm: DQN. The algorithm integrates deep neural networks into the Q-learning framework to approximate the Q-value function, thereby enabling efficient modeling of high-dimensional continuous state spaces. Moreover, DQN introduces the target network and experience replay mechanisms to mitigate sample correlation and target value instability, which significantly enhances training stability and convergence performance.
The eq.~\eqref{eq4.16} becomes:
\begin{equation}
	\label{DQN}
	 Q(s_k,a_k;\theta)\longleftarrow Q(s_k,a_k;\theta)+\alpha[r_{k}+\beta \max_{a \in \mathcal{A}(s_{k+1})} \hat{Q}(s_{k+1},a;\hat{\theta})-Q(s_k,a_k;\theta)],
\end{equation}
where the change in weights $\Delta \theta$ is given by:
\begin{equation}
	\label{DQN-Delta}
	\Delta \theta =\alpha[r_{k}+\beta \max_{a}\hat{Q}(s_{k+1},a;\hat{\theta})-{Q}(s_k,a_k;\theta)]\cdot\bigtriangledown  _{\theta }{Q}(s_k,a_k;\theta).
\end{equation}

Although DQN demonstrates significant advantages in handling high-dimensional tasks, it may still overestimate the values of optimal actions in complex state spaces with large action sets, potentially leading to unstable policies and difficulties in convergence. 
In the stochastic environment considered in our study,the final NPV of the project heavily depends on the long-term effects of the entire scheduling process. To effectively mitigate the bias caused by Q-value overestimation, we adopt DDQN algorithm as an enhancement to the standard DQN, thereby improving the stability of policy learning and the reliability of decision outcomes. The eq.~\eqref{DQN} is modified as follows:
\begin{equation}
	\label{DDQN}
	Q(s_k, a_k; \theta) \longleftarrow Q(s_k, a_k; \theta) 
	+ \alpha \Big[
	r_{k} 
	+ \beta \, \hat{Q}\big(s_{k+1}, \arg\max_{a \in \mathcal{A}(s_{k+1})} Q(s_{k+1}, a; \theta); \hat{\theta}\big) 
	- Q(s_k, a_k; \theta)
	\Big],
\end{equation}
where the temporal-difference (TD) error $\delta_k$ is defined as:
\begin{equation}
	\label{DDQN-TD}
	\delta_k = r_{k} + \beta \, \hat{Q}\big(s_{k+1}, \arg\max_{a \in \mathcal{A}(s_{k+1})} Q(s_{k+1}, a; \theta); \hat{\theta}\big) - Q(s_k, a_k; \theta),
\end{equation}
and the corresponding weight $\Delta \theta$ update is given by:
\begin{equation}
	\label{DDQN-Delta}
	\Delta \theta = \alpha \, \delta_k \, \nabla_{\theta} Q(s_k, a_k; \theta).
\end{equation}

\begin{figure}[H]
	\centering
	\begin{tikzpicture}[font=\sffamily]
		
		\node[circle, draw, thick, minimum size=1.9cm] (environment) {Environment};
		
		

		\node[draw, thick, rounded corners, minimum width=2.5cm, minimum height=1.5cm, left=2cm of environment] (q_predict) {Q Prediction};
	
		\node[draw, thick, rounded corners, minimum width=2.5cm, minimum height=1.5cm, left=5cm of q_predict] (q_target) {Q Target};
		
		\node[draw, thick, rounded corners, minimum width=5cm, minimum height=1cm, above=2cm of q_target, xshift=2cm] (loss) {Mean Squared Error loss};
		
		\node[draw, thick, rounded corners, minimum width=4cm, minimum height=1.5cm, below=1.5cm of q_target, xshift=3cm, align=center,text width=2cm] (replay_buffer) {Replay Buffer \par $(s_k, a_k, r_k, s_{k+1})$};
		
		\draw[->, thick] (environment.south) |- node[midway, above, xshift=-2cm] {\large  $s_k, a_k, r_k, s_{k+1}$} (replay_buffer.east);

		\draw[->, thick] (replay_buffer.north) |- ++(0,0.5) -| node[midway, above, xshift=0.5cm] {\large $s_k$}(q_predict.south);
		\draw[->, thick] (replay_buffer.north) |- ++(0,0.5) -| node[midway, above, xshift=-0.5cm] {\large $s_{k+1}$}(q_target.south);
		\draw[->, thick] (replay_buffer.west) -| ++(-3,0) |- node[midway, above, yshift=-6.3cm, xshift=1cm] {\large $r_k$}(loss.west);

		\draw[->, thick] (q_predict.east)  |- node[midway, below, xshift=1cm] {\large $a_k$} (environment.west);
		
		\draw[->, thick] ([yshift=-5pt]q_predict.west) |- 
		node[midway, below, xshift=-2.5cm] 
		{\scriptsize \shortstack{Every $C$ timesteps:\\ $\hat{\theta} \gets \theta$}}
		([yshift=-5pt]q_target.east);
		\draw[->, thick, dashed] ([yshift=5pt]q_predict.west) |- 
		node[midway, above, xshift=-2.5cm] 
		{\scriptsize \shortstack{$a_{k+1}=\arg\max_{a\in\mathcal{A}(s_{k+1})} Q(s_{k+1},a;\theta)$}}
		([yshift=5pt]q_target.east);
		
		\draw[->, thick] ([xshift=-8pt]q_predict.north) |- node[midway, left, yshift=-1cm] {\large $Q(s_{k},a_k;\theta)$} ([yshift=-8pt]loss.east);

		\draw[->, thick] (q_target.north) |- ++(0,1.2) -| node[midway, above, yshift=-1cm] {\large $\hat{Q}(s_{k+1},a_{k+1};\hat{\theta})$} (loss.south);
		
		\draw[->, thick, dashed] ([yshift=8pt]loss.east) -| node[midway, right, yshift=-1.5cm] {\large $\delta_k=y_k-Q(s_k,a_k;\theta)$} ([xshift=8pt]q_predict.north);
		
	\end{tikzpicture}
	\caption{Workflow of DDQN}
	\label{fig:workflow}
\end{figure}
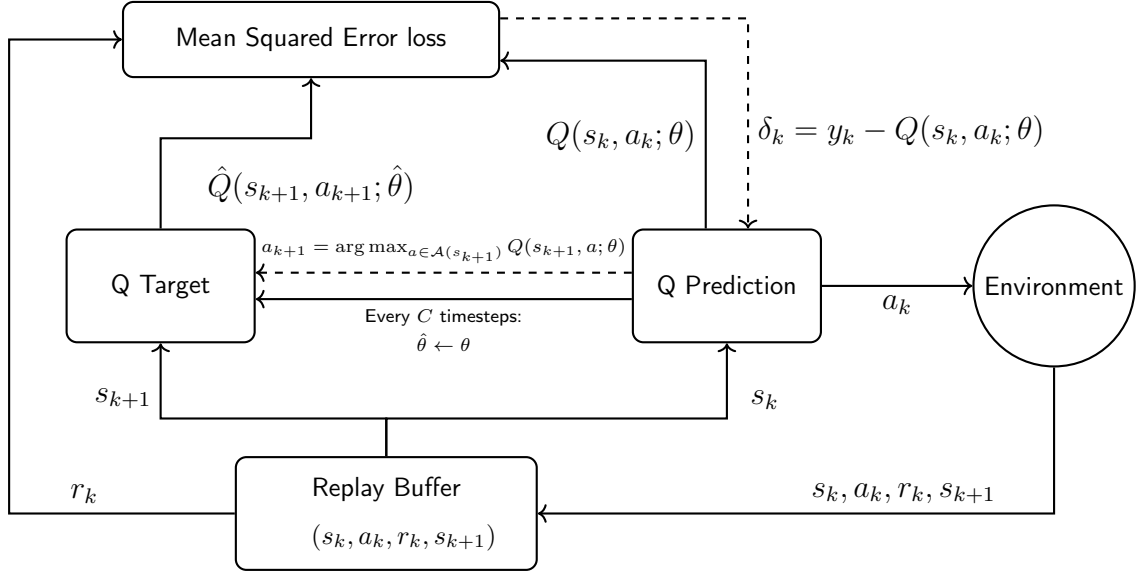

DDQN is an unsupervised reinforcement learning model based on the MDP. Its primary components include the online Q-network, the target network, and the replay buffer. The detailed workflow is illustrated in Figure \ref{fig:workflow}.

The replay buffer is a memory pool with a fixed capacity, used to store experiences generated during the agent-environment interactions, i.e., tuples$(s_k, a_k, r_k, s_{k+1})$, where $s_k$ denotes the current state, $a_k$ the action taken in state $s_k$, $r_k$ the immediate reward received after executing action $a_k$, and $s_{k+1}$ the subsequent state. The state representation in the MDP model has been detailed in Section~\ref{sec:Markov Decision Process(MDP)}. Meanwhile, the immediate reward $r_k$ is defined as the NPV obtained by executing a given activity $a_j$.

\begin{figure}[H]
	\centering
	\begin{tikzpicture}
		
		\tikzstyle{neuron}=[circle, draw=black, minimum size=20pt, inner sep=0pt]
		\tikzstyle{input neuron}=[neuron, fill=green!30]
		\tikzstyle{hidden neuron}=[neuron, fill=blue!30]
		\tikzstyle{output neuron}=[neuron, fill=red!30]
		\tikzstyle{annot} = [ text centered]
		
		\foreach \i in {1,...,4}
		\node[input neuron] (I-\i) at (0,{2.35 - 1.2*\i}) {};
		
		\node at (-2.5, 1.15) {current time: $t$};  
		\node at (-2.5, 0) {activities status: $x$};     
		\node at (-2.5, -1.2) {activities start time: $\varphi$};
		\node at (-2.5, -2.4) {activities duration: $\psi$};
		\draw [decorate,decoration={brace,amplitude=10pt,mirror,raise=2pt}] (-4.5, 1.15) -- (-4.5, -2.4) node [black,midway,xshift=-18pt] {$S$};
		
		\node[hidden neuron] (H1-1) at (2.5, 2) {};
		\node[hidden neuron] (H1-2) at (2.5, 1) {};
		\node at (2.5, 0) {\vdots}; 
		\node[hidden neuron] (H1-3) at (2.5, -1) {};
		\node[hidden neuron] (H1-4) at (2.5, -2) {};
		\node[hidden neuron] (H1-5) at (2.5, -3) {};
		
		\node[hidden neuron] (H2-1) at (4, 2) {};
		\node[hidden neuron] (H2-2) at (4, 1) {};
		\node at (4, 0) {\vdots}; 
		\node[hidden neuron] (H2-3) at (4, -1) {};
		\node[hidden neuron] (H2-4) at (4, -2) {};
		\node[hidden neuron] (H2-5) at (4, -3) {};
		
		\node[hidden neuron] (H3-1) at (5.5, 2) {};
		\node[hidden neuron] (H3-2) at (5.5, 1) {};
		\node at (5.5, 0) {\vdots}; 
		\node[hidden neuron] (H3-3) at (5.5, -1) {};
		\node[hidden neuron] (H3-4) at (5.5, -2) {};
		\node[hidden neuron] (H3-5) at (5.5, -3) {};
		
		\node[output neuron] (O-1) at (8, 2) {};
		\node[output neuron] (O-2) at (8, 1) {};
		\node at (8, 0) {\vdots}; 
		\node[output neuron] (O-3) at (8, -1) {};
		\node[output neuron] (O-4) at (8, -2) {};
		\node[output neuron] (O-5) at (8, -3) {};
		
		\node at (9.5, 2) {$Q(s,a_0)$};  
		\node at (9.5, 1) {$Q(s,a_1)$};  
		\node at (9.5, -1) {$Q(s,a_{n-1})$}; 
		\node at (9.5, -2) {$Q(s,a_{n})$}; 
		\node at (9.5, -3) {$Q(s,a_{n+1})$};  
		
		\foreach \i in {1,...,4}
		\foreach \j in {1,2,3,4,5}
		\draw[->] (I-\i) -- (H1-\j);
		
		\foreach \i in {1,2,3,4,5}
		\foreach \j in {1,2,3,4,5}
		\draw[->] (H1-\i) -- (H2-\j);
		
		\foreach \i in {1,2,3,4,5}
		\foreach \j in {1,2,3,4,5}
		\draw[->] (H2-\i) -- (H3-\j);
		
		\foreach \i in {1,2,3,4,5}
		\foreach \j in {1,2,3,4,5}
		\draw[->] (H3-\i) -- (O-\j);
		
		\node[annot] at (0, -4) {input layer};
		\node[annot] at (4, -4) {hidden layer};
		\node[annot] at (8, -4) {output layer};
		
	\end{tikzpicture}
	\caption{Structure diagram of DDQN}
	\label{fig:Structure diagram}
\end{figure}
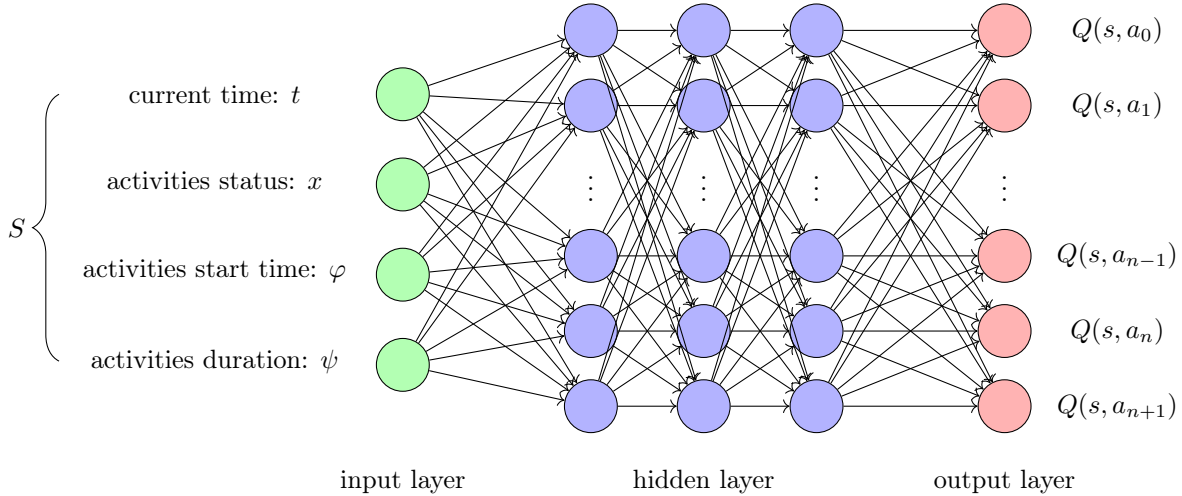

Both the online Q-network and the target network are implemented as deep Q-networks with identical architectures and initial parameters, designed to estimate the state-action value function ($Q(s, a)$). As illustrated in Figure~\ref{fig:Structure diagram}, the deep Q-networks consists of a five-layer fully connected network, including an input layer, multiple hidden layers, and an output layer. Nodes within the same layer are not interconnected, while nodes across different layers are fully connected. 
The number of units in the input layer is determined by the dimensionality of the state space($n_\text{input}=dim(s)$), whereas the number of units in the output layer corresponds to the number of project activities ($n_\text{output}=a_{n+1}$), with each output unit representing the Q-value of a feasible action. The number of neurons in the hidden layers determines the model complexity; in our study, three hidden layers are used with 256, 512, and 256 neurons respectively. All hidden layers employ the Rectified Linear Unit (ReLU) activation function to enhance nonlinear representation capability.

The key difference between the two deep Q-networks lies in their roles: the online network is used for training, while the target Q-network remains fixed over short intervals to provide stable target values.

As illustrated in Figure~\ref{fig:workflow}, the agent first extracts the overall project state information from the environment and stores it in a replay buffer to break the temporal correlation among samples and enhance training stability. During training, the agent randomly samples a minibatch of experiences from the replay buffer and feeds them into two neural networks. The online network receives the current state $s_k$ and outputs the Q-values corresponding to all feasible actions in the decision set $\mathcal{A}(s_k)$; the optimal action is then determined as $a_k = \arg\max_{a\in\mathcal{A}(s_k)} Q(s_k,a;\theta)$. If the selected action $a_0$ represents “no activity starts,” it indicates that no new activities are initiated at the current decision point. The environment then interacts based on the chosen action $a_k$, returning the immediate reward $r_k$ and the next state $s_{k+1}$. Subsequently, the online network selects the next action in the new state according to $a_{k+1} = \arg\max_{a\in\mathcal{A}(s_{k+1})} Q(s_{k+1},a;\theta)$, while the target network computes the TD target using the corresponding Q-value as $y_k = r_k + \beta \hat{Q}(s_{k+1},a_{k+1};\hat{\theta})$. This mechanism separates action selection from value evaluation, effectively mitigating the overestimation bias in Q-values. The online network parameters are updated by minimizing the TD error $\delta_k$. To ensure training stability, the target network parameters are periodically synchronized with the online network every fixed interval $C$, i.e., $\hat{\theta}\leftarrow\theta$.

The optimization objective of our study is to maximize the total expected NPV of the entire project using the DDQN algorithm. To this end, We set up a Python-based agent to facilitate interaction with the environment and iteratively update the Q-values. The primary training process is illustrated in Algorithm~\ref{alg:dqn}.
	
\begin{algorithm}[H]
	\centering
	\caption{DDQN Algorithm Based on MDP}
	\label{alg:dqn}
	\begin{algorithmic}
		\State \textbf{Step 1} Initialize the experience replay buffer $D$ with capacity $M$ and batch size $B$;
		\State \textbf{Step 2} Randomly initialize online Q-network parameters $\theta$ and target Q-network parameters $\hat{\theta} \gets \theta$;
		\For {episode = 1 to $N$}
		\State \quad  Initialize environment state $s_0$;  
		\State \quad  Total reward $R \gets 0$;
		\For{k = 1 to $T$} 
		\State \quad  Select action $a_k$ using $\varepsilon$-greedy policy;
		\State \quad  Execute action $a_k$, observe reward $r_k$ and next state $s_{k+1}$; 
		\State \quad  Store experience $(s_k, a_k, r_k, s_{k+1})$ in $D$; 
		\If{\(\textbar D \textbar \ge B\)}
		    \State \quad Randomly sample a batch of $B$ experiences $\{(s_\ell, a_\ell, r_\ell, s_{\ell+1})\}_{\ell=1}^B$ from $D$;
		    \State \quad Select next action using online network: $a_{\ell+1}^{\max} = \arg\max_a Q(s_{\ell+1}, a; \theta)$;
		    \State \quad Compute target Q-value using target network:
		      \begin{align*}
		      	y_\ell =
		      	\begin{cases}
		      		r_\ell & \text{if } s_{\ell+1} \text{ is done} \\
		      		r_\ell + \beta Q(s_{\ell+1}, a_{\ell+1}^{\max}; \hat{\theta}) & \text{otherwise}
		      	\end{cases};
		      \end{align*}
		    \State \quad Update online Q-network parameters $\theta$ by minimizing loss: $L = \frac{1}{B} \sum_\ell \big(y_\ell - Q(s_\ell, a_\ell; \theta)\big)^2$;
		    \State \quad Perform gradient descent step: $\theta \gets \theta - \alpha \nabla_\theta L$;
		\EndIf
		\State \quad Accumulate the rewards: $R \gets R + r_k$;
		\State \quad Every $C$ time steps update target network: $\hat{\theta} \gets \theta$;
		\EndFor 
		\State \quad Store or log total episode reward $R$.
	\EndFor 
	\end{algorithmic}
\end{algorithm}

\begin{example}
	\begin{figure}[h!]
		\centering
		\begin{tikzpicture}[>=stealth, node distance=2.2cm, font=\small]
			
			\node[circle, draw, minimum size=0.9cm] (0) at (0,0) {0};
			\node[above=0.1cm of 0] {0};
			\node[below=0.1cm of 0] {0};
			
			\node[circle, draw, minimum size=0.9cm] (1) [right=of 0] {1};
			\node[above=0.1cm of 1] {-90};
			\node[below=0.1cm of 1] {1};
			
			\node[circle, draw, minimum size=0.9cm] (2) [above right=1.5cm and 2cm of 1] {2};
			\node[above=0.1cm of 2] {-5500};
			\node[below=0.1cm of 2] {5};
			
			\node[circle, draw, minimum size=0.9cm] (3) [below right=1.5cm and 2cm of 1] {3};
			\node[above=0.1cm of 3] {-90};
			\node[below=0.1cm of 3] {\{1,10\}};
			
			\node[circle, draw, minimum size=0.9cm] (4) [right=of 1, xshift=2.2cm] {4};
			\node[above=0.1cm of 4] {+10000};
			\node[below=0.1cm of 4] {0};
			
			\node[circle, draw, minimum size=0.9cm] (i) [right=2cm of 4] {$i$};
			\node[above=0.1cm of i] {$\tilde{c}_i$};
			\node[below=0.1cm of i] {$\tilde{d}_i$};
			
			\draw[->] (0) -- (1) node[midway, above]{}; 
			\draw[->] (1) -- (2) node[midway, above left]{}; 
			\draw[->] (1) -- (3) node[midway, below left]{}; 
			\draw[->] (2) -- (4) node[midway, above right]{};
			\draw[->] (3) -- (4) node[midway, below right]{}; 
			
		\end{tikzpicture}
		\caption{The network of Example 1.}
		\label{fig:example}
	\end{figure}
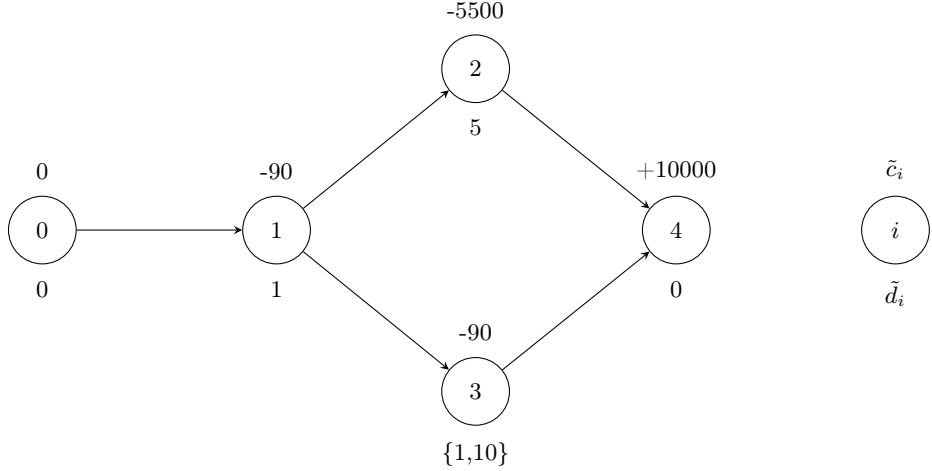
\end{example}

Figure~\ref{fig:example} illustrates an example project consisting of five activities under two possible scenarios, where $c^{1} = c^{2} = (0, -90, -5500, -90, +10000)$, $d^{1} = (0, 1, 5, 1, 0)$, and $d^{2} = (0, 1, 5, 10, 0)$. Each scenario occurs with an equal probability of $0.5$. The discount factor is set to $\beta = 0.9$, and the project’s maximum deadline is $\delta = 40$.

Under the assumption of perfect information, that is, when the decision-maker is aware of which scenario will actually occur, the optimal policy $\pi^{*}$ can be determined explicitly. 
In scenario~1, the optimal decision is to execute activities 0, 1, and 2 as early as possible ($t_{0}^{1} = t_{1}^{1} = 0$, $t_{2}^{1} = 1$), delay activity 3 by four time steps ($t_{3}^{1} = 5$), and then perform activity 4 once activities 2 and 3 are completed ($t_{4}^{1} = 6$), yielding a net present value of 
$\text{NPV}_1 = -90 \cdot 0.9^0 - 90 \cdot 0.9^1 - 5500 \cdot 0.9^5 + 10000 \cdot 0.9^6 = 221.80.$ 
For scenario 2, the optimal strategy $\Pi^*$ is to execute activities 0 and 1 first ($t_0^2 = t_1^2 = 0$), prioritize activity 3 ($t_3^2 = 1$), delay activity 2 by five time steps ($t_2^2 = 6$), and finally execute activity 4 ($t_4^2 = 11$), resulting in 
$\text{NPV}_2 = -90 \cdot 0.9^0 - 90 \cdot 0.9^6 - 5500 \cdot 0.9^1 + 10000 \cdot 0.9^{11} = 44.18.$
Hence, the expected NPV under perfect information is $\text{ENPV} = 0.5 \cdot \text{NPV}_1 + 0.5 \cdot \text{NPV}_2 = 132.72.$

In practical decision-making, however, the realized scenario is uncertain and cannot be known in advance.  
To address this challenge, we employ a DDQN algorithm that adaptively selects near-optimal actions in response to the dynamic evolution of the project state, thereby approximating the optimal scheduling policy under uncertainty.
Specifically, in scenario~1, the DDQN policy executes activities 0, 1, and 2 early ($t_{0}^{1} = t_{1}^{1} = 0$, $t_{2}^{1} = 1$), advances activity~3 slightly to time step~4 ($t_{3}^{1} = 4$), and completes activity 4 at time step 6 ($t_{4}^{1} = 6$). The resulting NPV is $\text{NPV}_1 = -90 \times 0.9^{0} - 90 \times 0.9^{1} - 5500 \times 0.9^{4} + 10000 \times 0.9^{6} = 215.36$. For Scenario~2, the DDQN policy coincides with the optimal strategy under perfect information ($t_{0}^{2} = t_{1}^{2} = 0$, $t_{3}^{2} = 1$, $t_{2}^{2} = 6$, $t_{4}^{2} = 11$), yielding the same $\text{NPV}_2 = 44.18$. Therefore, the expected NPV obtained under DDQN is $\mathrm{ENPV} = 0.5 \times \mathrm{NPV}_{1} + 0.5 \times \mathrm{NPV}_{2} = 129.77$.

The comparison reveals that, in the absence of prior scenario information, the expected net present value achieved by DDQN decreases only slightly—from $132.72$ to $129.77$, a reduction of approximately $2.2\%$. This indicates that although DDQN does not possess perfect information, it can still effectively approximate the optimal decision through adaptive learning. More importantly, DDQN substantially enhances the robustness of the scheduling policy across different scenarios, maintaining stable performance under uncertainty. Furthermore, since its decision-making process does not rely on pre-specified scenario knowledge, the proposed approach exhibits high practical applicability and provides valuable insights for dynamic optimization in real-world project management.

\section{Computational results}
\label{sec:computationalResults}
In this section, we conduct validation experiments on the constructed instances to evaluate the performance of the proposed DDQN algorithm. Subsection~\ref{sec:parameterSettings} presents the parameter settings of the DDQN network structure, while Subsection~\ref{sec:dateset} describes the procedure for dataset generation.
Subsection~\ref{sec:baselineMethods} provides a concise overview of the fundamental concepts underlying the comparative algorithms.
The evaluation metrics for method comparison are introduced in Subsection~\ref{sec:evaluationMetric}. In Subsection~\ref{sec:resultsAnalysis}, the proposed approach is compared against metaheuristic algorithms. 

\subsection{Parameter Settings}
\label{sec:parameterSettings}
All experiments were conducted on the same workstation equipped with a 13th Gen Intel\textsuperscript{\textregistered} Core\texttrademark~i7-13700H processor (2.40~GHz). The algorithms used in the experiments were implemented in Python~3.12 and the neural network models were deployed using the PyTorch~2.5.1 framework. Table~\ref{tab:network-parameters} summarizes the parameter settings of the DDQN algorithm's network architecture.

\begin{table}[h] 
	\centering
	\caption{Network Parameters}
	\label{tab:network-parameters}
	\begin{tabular}{@{}c c c@{}} 
		\toprule
		\multirow{5}{*}{Network architecture} 
		& Hidden layers & 3 \\
		& Hidden units  & 256, 512 \\
		& Activation    & Relu \\
		& Input size    & State space \\
		& Output size   & Number of activities \\
		\midrule
		\multirow{4}{*}{Training hyperparameters} 
		& Learning rate & 1e-5 \\
		& Discount factor & 0.9 \\
		& Loss function   & MSE \\
		& optimizer       & Adam \\
		\midrule
		\multirow{2}{*}{Replay buffer} 
		& Buffer size & 50000 \\
		& Batch size  & 256 \\
		\midrule
		\multirow{3}{*}{$\varepsilon$-greedy strategy} 
		& Epsilon start & 1.0 \\
		& Epsilon end   & 0.01 \\
		\midrule
		\multirow{1}{*}{Target network} 
		& Target update frequency & 1000 \\                      
		\bottomrule
	\end{tabular}
\end{table}

\subsection{Dateset}
\label{sec:dateset}
In this section, we describe the generation of two sets of datasets. Due to the stochasticity of activity durations and cash flows, we employ a Monte Carlo simulation approach to determine the project deadlines, denoted as $\delta$, for both sets of instances. Specifically, for each task, random completion times are generated based on its duration distribution, and the simulation is repeated 10,000 times to cover various possible scenarios. The project completion date is then determined through statistical analysis of the simulation results at a 90\% confidence level. The discount factor is set to $\beta = 0.9$.

The first set of instances, denoted as $\Omega_1$, is constructed following the framework proposed by \cite{rostami2024}. 
It generates a limited number of scenarios, for each activity $j$, its duration is randomly discrete sampled from a uniform distribution over $(1,10)$. The activity cash flow is computed as $\tilde{c}_j = c_j^F + \tilde{d}_j \cdot c_j^V + g_j$ (see Section~\ref{sec:problemStatement}), where the fixed cost $c_j^F$ is drawn from a uniform distribution over $(-10,-1)$, the variable cost $c_j^V$ is drawn from a uniform distribution over $(-10,-1)$, and the benefit $g_j$ is drawn from a uniform distribution over $(0,100)$. 
Unlike \cite{rostami2024}, where cash flows are directly sampled uniformly from the interval $(-100,100)$, this generation approach more realistically reflects how cash flows vary with activity durations. It also increases data diversity and complexity, thereby facilitating a more comprehensive evaluation of scheduling policy performance.

Furthermore, the precedence network is constructed using randomly generated arcs as in \cite{rostami2024}. For each pair $i,j \in V$, if $(j,i) \notin E$, the arc $(i,j)$ and its transitive arcs are added to $E$ with a probability of 0.2. Using this procedure, for $(n-2) \in \{5,10,15,20,25,30\}$ and $\Sigma \in \{2,5,10\}$, 10 instances are generated for each parameter setting.

The second set of instances, denoted as $\Omega_{2}$, imposes no restriction on the number of scenarios and is primarily designed to simulate online or dynamic scheduling environments. 
Unlike the first set of instances, no explicit scenarios are generated. 
For each activity $j$, only the corresponding fixed cost $c_{j}^{F}$, variable cost $c_{j}^{V}$, and revenue $g_{j}$ are pre-generated, while the activity duration $\tilde{d}_{j}$ is not predetermined during instance generation. Since the cash flow $\tilde{c}_{j} = c_{j}^{F} + \tilde{d}_{j} \cdot c_{j}^{V} + g_{j}$ depends on the realized duration $\tilde{d}_{j}$, it cannot be generated in advance but is dynamically determined during the scheduling execution process. 
For different activity sizes $(n - 2) \in \{5, 7, 10, 12, 15, 20, 25, 30\}$, 10 random instances are generated for each configuration, in order to evaluate the adaptability of the scheduling policy under conditions where future information is unavailable.

\subsection{Baseline Methods}
\label{sec:baselineMethods}
We select two representative strategies from \cite{rostami2024} as baselines to evaluate the performance of the proposed DDQN algorithm: the rigid policy and the dynamic policy. The following sections briefly summarize the key characteristics and operational principles of each baseline.

\subsubsection{Rigid Policy}
\label{sec:rigidPolicy}
A rigid policy is a type of static project scheduling strategy characterized by assigning a fixed starting time $t_j$ for each activity $j \in V$ prior to project execution, which remains unchanged across all scenarios $\sigma \in \Sigma$. In other words, a rigid policy does not adjust the activity start times in response to the realization of uncertain factors.
Since the activity durations $d_j^\sigma$ may differ across scenarios, to ensure feasibility under all scenarios, a rigid policy must satisfy precedence constraints under the worst-case durations. Its mathematical formulation can be expressed as follows:
\begin{subequations}
	\begin{align}
		\max \quad & \sum_{\sigma \in \Sigma} p_\sigma \sum_{j \in V} c_j^\sigma \beta^{t_j} 
		\label{rigid} \\[1ex]
		\text{s.t.} \quad & t_j \ge t_i + d_i^\sigma, \ \  \forall (i,j) \in E, \forall \sigma \in \Sigma 
		\label{rigid-1} \\[1ex]
		& t_{n+1} \le \delta
		\label{rigid-2} \\[0.5ex]
		& t_0 = 0
		\label{rigid-3} \\[0.5ex]
		& t_j \ge 0, \ \ t_j \in \mathbb{Z}, \forall\ j\in\ V.
		\label{rigid-4}
	\end{align}
\end{subequations}

The constraints ensure that each activity’s fixed start time $t_j$ satisfies precedence relations and project duration limits in all possible scenarios, thereby guaranteeing non-anticipativity. However, such a policy is generally conservative, typically scheduling activities based on their maximum durations, which may result in a lower expected net present value.

\subsubsection{Dynamic Policy}
\label{sec:dynamicPolicy}
A dynamic policy initially stipulates that the start time of each activity be postponed until after the completion of its predecessor activities by a lag time that is independent of the scenario, specifically the minimum delay $l_{ij}$, which represents the minimal interval between the completion of activity $i$ and the initiation of activity $j$. 
Simultaneously, the policy allows the project to adjust the start times of subsequent activities in real time based on realized uncertainties during execution, such as the actual durations of completed activities. For each scenario $\sigma \in \Sigma$, the policy $\pi$ maps it to the corresponding set of start times $\{t_{j}^{\sigma}\}_{j \in V}$.

The optimization model under a dynamic policy can be formulated as:
\begin{subequations}
	\begin{align}
		\max \quad & \sum_{\sigma \in \Sigma} p_\sigma \sum_{j \in V} c_j^\sigma \, \beta^{t_j^\sigma}
		\label{DYN} \\[1ex]
		\text{s.t.} \quad & t_j^\sigma \ge t_i^\sigma + d_i^\sigma + l_{ij}, \quad \forall (i,j) \in E, \ \forall \sigma \in \Sigma 
		\label{DYN-1} \\[1ex]
		& l_{ij} \ge 0, \quad \forall (i,j) \in E
		\label{DYN-2} \\[0.5ex]
		& t_{n+1}^\sigma \le \delta, \quad \forall \sigma \in \Sigma
		\label{DYN-3} \\[0.5ex]
		& t_0^\sigma = 0, \quad \forall \sigma \in \Sigma
		\label{DYN-4} \\[0.5ex]
		& t_j^\sigma \ge 0, \quad t_j^\sigma \in \mathbb{Z}, \quad \forall j \in V, \ \forall \sigma \in \Sigma
		\label{DYN-5}
	\end{align}
\end{subequations}

Constraint~\eqref{DYN-1} stipulates that the start time of each activity $j$ must be greater than the completion time of its predecessor activity $i$, that is, the predecessor's start time plus its duration and the minimum lag. 
The decision depends only on the information of activities that have already been completed, thereby satisfying the non-anticipativity requirement.
Constraint~\eqref{DYN-2} specifies that the minimum lag between activities is non-negative. 
Constraints~\eqref{DYN-3}--\eqref{DYN-5} are identical to those presented in Section~\ref{sec:problemStatement} and are therefore omitted here.

At each decision time $t$, a heuristic approach is employed that considers only the currently eligible activity $j$. The decision is guided by comparing the upper bound of the expected value under perfect information (EV$|$PI) for two alternatives: immediate start versus delayed start. Specifically, 
\begin{equation}
	\text{EV|PI}(t,j) = \sum_{\sigma \in \Sigma} p_\sigma \, N_{\max}^{*\sigma},
\end{equation}
where $N_{\max}^{*\sigma}$ denotes the deterministic optimal NPV for a given scenario $\sigma$. For example, if $\text{EV$|$PI}(t,j) > \text{EV$|$PI}(t',j)$, the potential gain from starting activity $j$ at the current time is higher, and it should therefore be executed immediately; otherwise, it is postponed to the next decision time $t'$ to acquire more information before making the decision.

This heuristic computes decisions without enumerating all possible sequences, allowing rapid generation of high-quality approximate solutions for small to medium-sized problems. However, its performance heavily depends on the quality of the heuristic rules and accumulated experience. It lacks adaptive learning capability and cannot guarantee global optimality in complex or unknown environments. As the problem size 
$|V|$ or scenario space $|\Sigma|$ increases, constructing and tuning rules becomes increasingly costly, which may reduce computational efficiency and decision quality. In nonlinear, high-dimensional, or dynamic environments, fixed rules have limited adaptability and require frequent manual adjustments.

\subsection{Evaluation Metric}
\label{sec:evaluationMetric}
The objective of this study is to maximize the expected NPV. For the DDQN algorithm, we compute the average NPV over approximately 1,000 training runs to estimate its performance. To more intuitively assess the solution quality of different algorithms across various instances, we employ the relative percentage gap as the performance evaluation metric.
It is calculated as follows:
\begin{equation}
	\label{eq5.1}
	\text{Gap} = 100 \times \frac{N_{\max}^{*} - N_{\max}}{N_{\max}^{*}}
\end{equation}

where $N_{\max}$ denotes the expected NPV obtained by the DDQN algorithm, and $N_{\max}^{*}$ represents the optimal NPV under deterministic conditions with a finite set of scenarios. The lagger the expected NPV obtained by the algorithm is, the smaller the relative scheduling gap with respect to $N_{\max}^{*}$ becomes, and this indicates a higher solution quality.

\subsection{Results Analysis}
\label{sec:resultsAnalysis}
To evaluate the performance of the DDQN algorithm, its predicted outcomes are compared with the rigid policy and dynamic policy proposed by \cite{rostami2024}. 

\subsubsection{Experiment 1: Fixed Scenarios}
\label{sec:experiment 1}
Based on the training dataset $\Omega_{1}$, a set of fixed scenarios $\sigma \in \Sigma$ is defined, where each scenario $\sigma$ corresponds to a specific set of activity durations and cash flows $\{(d_{j}^{\sigma}, c_{j}^{\sigma})\}_{j \in V}$. During the training process, the agent randomly switches among these scenarios to learn a generalized scheduling policy that performs effectively across multiple scenario conditions.

Table~\ref{tab:Experiment 1} reports the average gap percentages of the Rigid, DYN, and DDQN algorithms, where Rigid denotes the rigid scheduling policy and DYN corresponds to the best dynamic policy obtained through heuristic approaches.

\begin{table}[h] 
	\centering
	\caption{Experiment 1}
	\label{tab:Experiment 1}
	\begin{tabularx}{\textwidth}{@{}Y Y Y Y Y@{}} 
		\toprule
		$n-2$ & $|\Sigma|$ & Rigid & DYN & DDQN \\ 
		\midrule
		      & 2  & 134.88 & 30.46 & 45.92 \\
		 5    & 5  & 92.60  & 20.14 & 35.74  \\
		      & 10 & 86.57  & 23.26 & 31.71  \\
		\midrule
		      & 2  & 114.21 & 15.44 & 32.71  \\
		 10   & 5  & 100.88 & 11.41 & 23.21 \\
		      & 10 & 91.35  & 9.67  & 19.01  \\
		\midrule
		      & 2  & 107.46 & 7.09 & 6.21 \\
		15    & 5  & 92.26  & 4.79 & 19.02 \\
		      & 10 & 91.42  & 5.30 & 9.14 \\
		\midrule
		      & 2  & 113.42 & 9.26 & 8.24 \\
		20    & 5  & 97.86  & 4.79 & 10.46 \\
		      & 10 & 102.18 & 6.33 & 8.99  \\
		\midrule
		      & 2  & 122.01 & 5.05 & 12.41  \\
		25    & 5  & 99.38  & 5.17 & 5.76 \\
		      & 10 & 95.16  & 6.43 & 4.66   \\
		\midrule 
		      & 2  & 106.62 & 3.20  & 4.30 \\
		30    & 5  & 95.15  & 4.67  & 3.95 \\
		      & 10 & 96.22  & 5.90  & 3.57    \\                  
		\bottomrule
	\end{tabularx}
\end{table}

As shown in table~\ref{tab:Experiment 1}, both the dynamic policy and the DDQN algorithm significantly outperform the rigid policy in overall performance, indicating that a state-feedback-based dynamic decision mechanism can effectively enhance scheduling quality. A further comparison reveals that for small activity sizes ($(n-2) \leq 10$), the dynamic policy performs better than DDQN. However, as the number of activities increases, the advantage of DDQN becomes more pronounced. For example, when $(n-2) = 15$ and $|\Sigma| = 2$, the gap of the dynamic policy is 7.09\%, whereas that of DDQN decreases to 6.21\%. When $(n-2) = 25$ and $|\Sigma| = 10$, the dynamic policy gap is 6.43\%, while DDQN further reduces it to 4.66\%. For $(n-2) = 30$ and $|\Sigma| = 10$, the dynamic policy achieves a gap of 5.90\%, whereas DDQN attains only 3.57\%.

Based on the above analysis, DDQN generally outperforms the dynamic policy in most scenarios, particularly when the number of activities is large or the environmental complexity is high. In such cases, DDQN can leverage the reinforcement learning mechanism to better capture the nonlinear relationships between state transitions and policy decisions, thereby achieving a higher expected net present value. However, in some small-scale or low-complexity scenarios, where the state space is relatively limited, the dynamic policy can fully exploit the current state and available information to make near-optimal decisions. 
In contrast, DDQN may suffer from insufficient training samples in small-scale problems, resulting in inferior performance. Overall, DDQN demonstrates superior generalization and long-term scheduling performance compared to the dynamic policy, exhibiting stronger capability in policy learning and environmental adaptation.

\subsubsection{Experiment 2: Stochastic Perturbations}
\label{sec:experiment 2}
The DDQN is trained based on the dataset $\Omega_{2}$, where each episode is regarded as a complete and independent project scenario $\sigma$. At each decision time $t_{k}$, the agent selects and executes the action $a_{k}$ corresponding to the maximum $Q$-value from the current feasible action set $\mathcal{A}(s_{k})$, that is, executing a specific activity $j$. The duration $\tilde{d}_{j}$ of the activity $j$ is randomly discrete sampled from a uniform distribution over $(1,10)$, based on which the associated cash flow $\tilde{c}_{j}$ and the immediate reward $r_{k}$ are determined. 
Since the durations of activities in each episode are independently generated during execution, and the cash flows of each activity are determined by their respective durations, the training process can be viewed as decision optimization under multiple randomly perturbed scenarios of similar projects. By continuously learning in these randomized environments, the agent progressively accumulates experience and thereby develops a scheduling policy that exhibits strong robustness under uncertainty.

Table~\ref{tab:Experiment 2} presents the average gap percentages of the Rigid, DYN, and DDQN algorithms. The DDQN agent is trained for a fixed number of 20,000 episodes for each instance.

As shown in the table, both the dynamic policy and DDQN continue to outperform the Rigid policy. Unlike the first experiment, DDQN consistently outperforms the dynamic policy across all instances, and its advantage becomes more pronounced as the number of activities increases. For example, when $(n-2) = 5$, the gap between the dynamic policy and DDQN is relatively small (9.21\% vs. 8.14\%), whereas for $(n-2) = 30$, the average gap of the dynamic policy rises to 57.03\%, while DDQN maintains a gap of 29.95\%. This improvement is attributed to the introduction of stochastic perturbations, which make future states unpredictable. As the number of activities increases, the state space grows exponentially, limiting the heuristic-based dynamic policy in fully evaluating all possible states. In contrast, DDQN can accumulate experience and optimize long-term rewards through reinforcement learning over a large number of stochastic training scenarios. 
This experiment further validates that the DDQN algorithm possesses stronger learning capability and generalization performance in complex environments.

\begin{table}[h] 
	\centering
	\caption{Experiment 2}
	\label{tab:Experiment 2}
	\begin{tabularx}{\textwidth}{@{}Y Y Y Y Y@{}} 
		\toprule
		$n-2$ & Rigid & DYN & DDQN \\ 
		\midrule
		5  & 14.93 & 9.21 & 8.14 \\
		7  & 37.14 & 27.78 & 18.48 \\
		10 & 69.62 & 36.56 & 25.65\\
		12 & 114.18 & 40.12 & 29.79\\ 
		15 & 131.12	& 44.70 & 32.24\\
		20 & 112.19	& 49.19 & 34.44 \\
		25 & 114.73 & 50.99	& 36.97 \\
		30 & 105.10 & 57.03 & 29.95\\                    
		\bottomrule
	\end{tabularx}
\end{table}

Based on the results of both experiments, it is evident that the global optimality of the dynamic policy cannot be guaranteed. As the number of candidate actions increases, the combinatorial explosion significantly raises computational complexity, limiting scalability in large-scale or highly dynamic uncertain environments. In contrast, DDQN approximates the Q-value function by training a neural network through reinforcement learning, allowing it to directly select actions that maximize long-term expected rewards in high-dimensional state spaces. This approach avoids the need to explicitly enumerate all possible action combinations and enables the policy to adapt in dynamic and stochastic environments, thereby increasing the likelihood of achieving near-global optimal solutions. However, DDQN entails higher training costs, reduced interpretability, and potentially unstable convergence. Therefore, in small-scale problems or scenarios requiring high interpretability, heuristic-based dynamic policies still hold significant advantages.

\subsection{Ablation experiment}
In this section, we conduct ablation experiments on the DDQN algorithm to examine the impact of its key components on performance. Three comparative schemes are designed: (i) DDQN, which employs the dual-network architecture for separate action selection and target value computation; (ii) DQN, the standard single-network version; and (iii) No Target, in which the target network is removed and the same network is used for both parameter updates and value estimation. All three methods are trained under identical environmental configurations and hyperparameter settings, and their average rewards are evaluated across training episodes in two task scenarios with $(n-2) = 10$ and $(n-2) = 30$.

\begin{figure}[htbp]
	\centering
	\begin{minipage}[b]{0.45\textwidth}
		\centering
		\includegraphics[width=\textwidth]{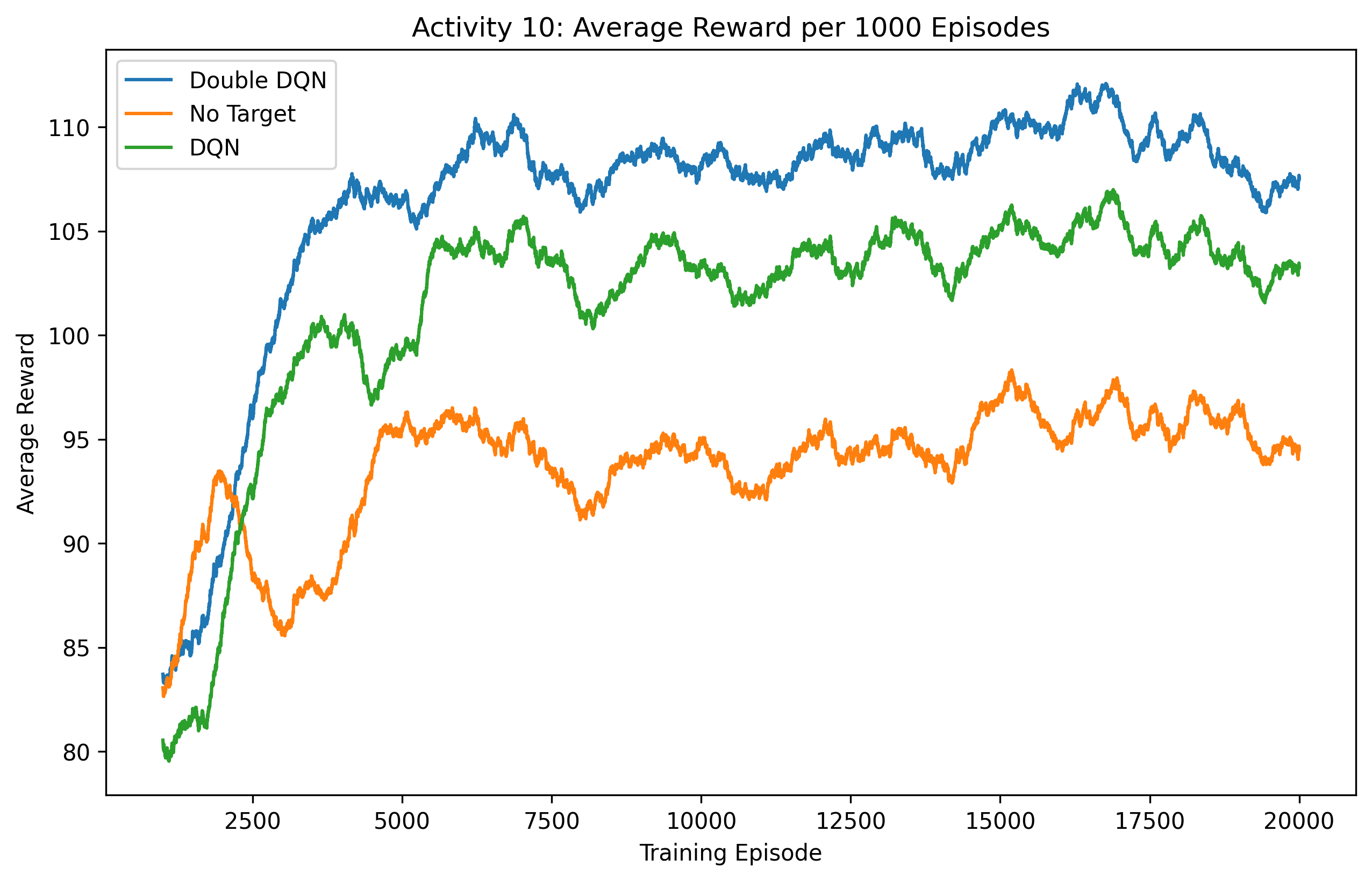}
	\end{minipage}
	\hspace{0.05\textwidth} 
	\begin{minipage}[b]{0.45\textwidth}
		\centering
		\includegraphics[width=\textwidth]{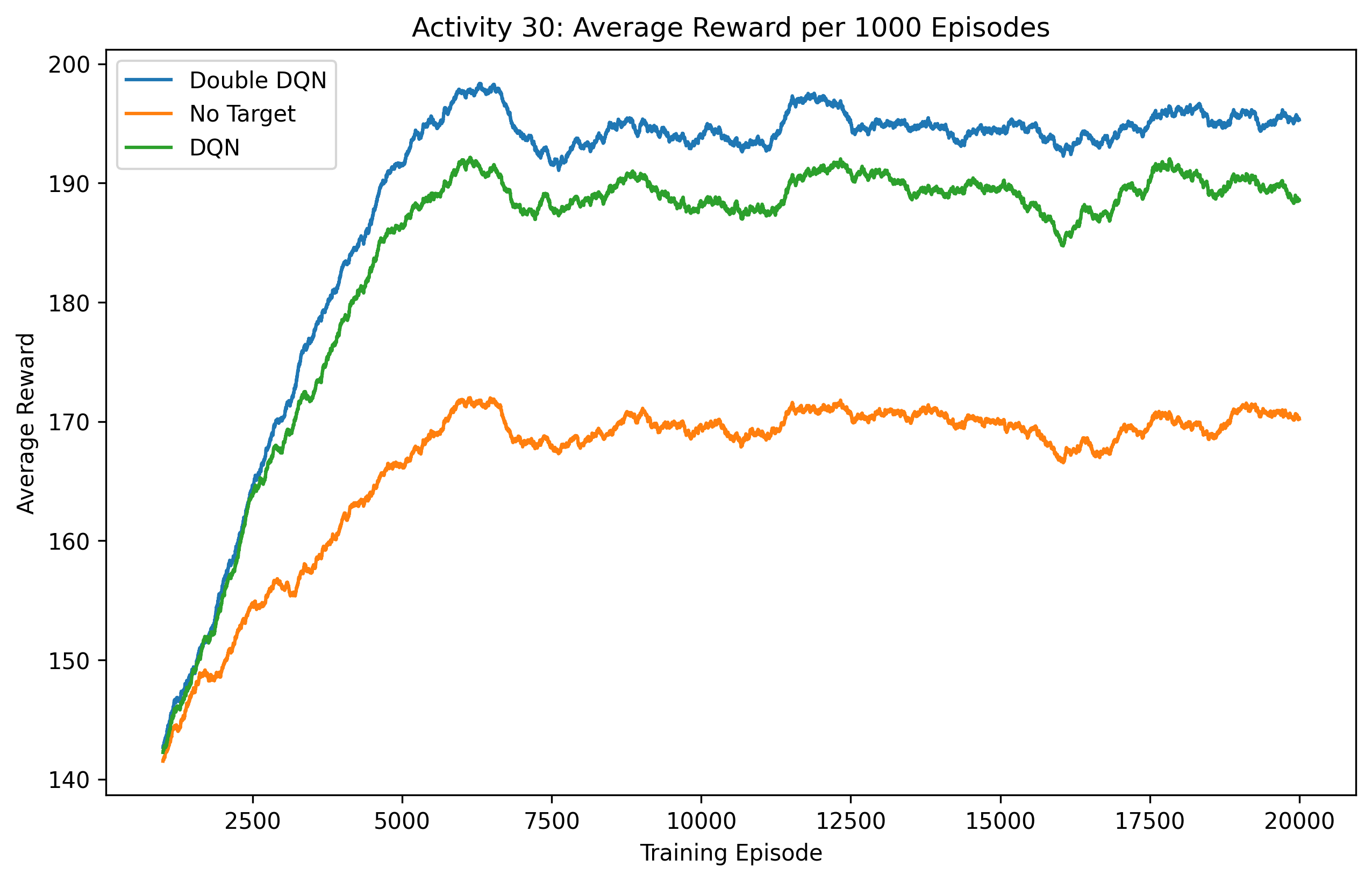}
	\end{minipage}
	\caption{Results of ablation experiments}
	\label{fig:ablation experiments}
\end{figure}

As shown in Figure~\ref{fig:ablation experiments}, the performance of DDQN is consistently superior to that of DQN and No Target across both tasks. Specifically, DDQN rapidly increases the reward in the early training stage (approximately 2,000–5,000 episodes) and maintains steady growth in the later stages. Although DQN achieves some improvement, it still exhibits slight overestimation, resulting in a lower final average reward. The model without a target network performs the worst, with highly fluctuating training curves and unstable convergence, highlighting the critical role of the target network in stabilizing the training process.

Overall, the results of the ablation experiments validate the effectiveness of the DDQN design. The dual-network architecture effectively mitigates the overestimation of action values, enabling the model to achieve higher and more stable rewards across different tasks. Additionally, the introduction of the target network substantially enhances training convergence and robustness. These findings demonstrate that DDQN outperforms the standard DQN in both structure and performance, providing a more reliable solution for stable decision-making in reinforcement learning.

\section{Conclusions}
\label{sec:conclusions}
We address the project’s expected NPV maximization problem, characterized by stochastic activity durations and cash flows under discrete scenarios, by formulating it as a discrete-time MDP and solving it with the DDQN algorithm. 
We evaluate the proposed approach through comparative experiments on multiple datasets against traditional heuristic methods. 
To elucidate the influence of critical components on algorithm performance, we conduct ablation studies to quantify the contributions of the dual-network architecture and the target network in enhancing model stability and optimization efficiency.

Experimental results demonstrate that the DDQN algorithm consistently outperforms the rigid policy across most scenarios. In Experiment~1, under relatively predictable future states, the dynamic policy slightly surpasses DDQN in small-scale problems, but DDQN exhibits superior computational efficiency and stability as problem size increases. Experiment~2 shows that with rising environmental stochasticity and less predictable future states, DDQN’s advantages become more pronounced, effectively handling complex, high-dimensional environments. 
Ablation studies further validate the DDQN architecture, indicating that the dual-network mechanism and target network substantially enhance convergence stability and overall performance.

In summary, the DDQN algorithm exhibits strong adaptability and robustness in large-scale or highly uncertain project scenarios, highlighting its potential for practical project management and decision optimization.
However, it requires extensive training samples and computational resources, is sensitive to hyperparameters, and may exhibit slow convergence or reduced policy stability in extremely high-dimensional or sparse-reward environments. Future research may incorporate continuous-action reinforcement learning and uncertainty modeling to improve stability, sample efficiency, and practical applicability.

\section*{Acknowledgements}
This work is supported by the Young Scientists Fund of the National Natural Science Foundation of China (grant number 72201209) and the Shanghai Pujiang Program (grant number 22PJC091). 

\appendix
\renewcommand{\thetable}{A\arabic{table}}
\setcounter{table}{0}

\section{Notations Used in the Model}

\begin{table}[h!]
	\centering
	\caption{Notation of project and activity}
	\small
	\begin{tabularx}{\textwidth}{>{\centering\arraybackslash}p{3cm}X >{\raggedright\arraybackslash}X}
		\hline
		\multicolumn{1}{c}{\textbf{Symbol}} & \multicolumn{1}{c}{\textbf{Description}} \\
		\hline
		$G=(V,E)$ & Single-mode network graph \\
		$i,j$ & Activity indices \\
		$V=\{0,1,\dots,n+1\}$ & Set of project activities \\
		$V'$ & Subset of activities with positive cash inflow \\
		$E$ & Precedence relations, $(i,j)\in E$ indicates $j$ starts after $i$ finishes \\
		$\tilde{d}_j$ & Duration of activity $j$ \\
		$\tilde{c}_j$ & Cash flow of activity $j$ \\
		$c_j^F$ & Fixed cost of activity $j$ \\
		$c_j^V$ & Unit variable cost of activity $j$ \\
		$g_j$ & Revenue from activity $j$ \\
		$\delta$ & Project deadline \\
		$\Pi$ & Set of all feasible policies \\
		$\pi$ & A single policy, $\pi \in \Pi$ \\
		$\pi^*$ & Optimal policy \\
		$\Sigma$ & Set of all possible scenarios \\
		$\sigma$ & A single scenario, $\sigma \in \Sigma$ \\
		$p_\sigma$ & Probability of scenario $\sigma$, $p_\sigma = 1/|\Sigma|$ \\
		$d^\sigma$ & Activity durations under scenario $\sigma$ \\
		$c^\sigma$ & Activity cash flows under scenario $\sigma$ \\
		$t_j^\sigma$ & Start time of activity $j$ under scenario $\sigma$ \\
		$d_j^\sigma$ & Duration of activity $j$ under scenario $\sigma$ \\
		$c_j^\sigma$ & Cash flow of activity $j$ under scenario $\sigma$ \\
		\hline
	\end{tabularx}
\end{table}

\begin{table}[H]
	\centering
	\caption{Notation of MDP and DDQN}
	\small
	\begin{tabularx}{\textwidth}{>{\centering\arraybackslash}p{3cm} >{\raggedright\arraybackslash}X}
		\hline
		\multicolumn{1}{c}{\textbf{Symbol}} & \multicolumn{1}{c}{\textbf{Description}} \\
		\hline
		$\mathcal{S}$ & The finite state space \\
		$\mathcal{A}$ & The finite action space \\
		$k$ & Current time step or decision stage \\
		$s_k$ & State space at decision stage $k$ \\
		$t_k$ & Current decision time \\
		$A_k$ & Set of activities in progress at time $k$ \\
		$C_k$ & Set of activities completed in $[t_k,t_{k+1})$ \\
		$U_k$ & Set of not-started activities at time $k$ \\
		$F_k$ & Set of completed activities at time $k$ \\
		$\boldsymbol{\varphi}_{k}$ & Vector of activity durations at time $k$ \\
		$\boldsymbol{\psi}_{k}$ & Vector of activity start times at time $k$ \\
		$\boldsymbol{x}_k$ & Activity state vector at time $k$ (0: not started, 1: in progress, 2: completed) \\
		$\mathcal{A}(s_k)$ & Feasible action set under state $s_k$ \\
		$a_k$ & Action executed at time $k$, $a_k \in A(s_k)$ \\
		$p(s_{k+1} \mid s_k, a_k)$ & Transition probability from state $s_k$ to $s_{k+1}$ under action $a_k$ \\
		$\pi(a_k \mid s_k)$ & Probability that policy $\pi$ selects action $a_k$ in state $s_k$ \\
		$r(s_k,a_k), r_k$ & Immediate reward from executing $a_k$ in state $s_k$ \\
		$V^\pi(s_k)$ & State value function following policy $\pi$ \\
		$Q^\pi(s_k,a_k)$ & State-action value function following policy $\pi$ \\
		$\alpha$ & Learning rate \\
		$y_k$ & TD target \\
		$\delta_k$ & TD error \\
		$C$ & Target network update frequency \\
		$\theta$ & Online network parameters \\
		$\hat{\theta}$ & Target network parameters \\
		\hline
	\end{tabularx}
\end{table}

\clearpage 
\bibliographystyle{abbrvnat}
\bibliography{reference}

\begin{thebibliography}{31}
\providecommand{\natexlab}[1]{#1}
\providecommand{\url}[1]{\texttt{#1}}
\expandafter\ifx\csname urlstyle\endcsname\relax
  \providecommand{\doi}[1]{doi: #1}\else
  \providecommand{\doi}{doi: \begingroup \urlstyle{rm}\Url}\fi

\bibitem[Asadujjaman et~al.(2021)Asadujjaman, Rahman, Chakrabortty, and
  Ryan]{asadujjaman2021}
M.~Asadujjaman, H.~F. Rahman, R.~K. Chakrabortty, and M.~J. Ryan.
\newblock Resource constrained project scheduling and material ordering problem
  with discounted cash flows.
\newblock \emph{Computers \& Industrial Engineering}, 158:\penalty0 107427,
  2021.

\bibitem[Avalos and Ortiz(2023)]{avalos2023}
S.~Avalos and J.~M. Ortiz.
\newblock Multivariate geostatistical simulation and deep q-learning to
  optimize mining decisions.
\newblock \emph{Mathematical Geosciences}, 55\penalty0 (5):\penalty0 673--692,
  2023.

\bibitem[Benati(2006)]{benati2006}
S.~Benati.
\newblock An optimization model for stochastic project networks with cash
  flows.
\newblock \emph{Computational Management Science}, 3:\penalty0 271--284, 2006.

\bibitem[Buss and Rosenblatt(1997)]{buss1997}
A.~H. Buss and M.~J. Rosenblatt.
\newblock Activity delay in stochastic project networks.
\newblock \emph{Operations Research}, 45\penalty0 (1):\penalty0 126--139, 1997.

\bibitem[Cai et~al.(2024)Cai, Bian, and Liu]{cai2024}
H.~Cai, Y.~Bian, and L.~Liu.
\newblock Deep reinforcement learning for solving resource constrained project
  scheduling problems with resource disruptions.
\newblock \emph{Robotics and Computer-Integrated Manufacturing}, 85:\penalty0
  102628, 2024.

\bibitem[Chen et~al.(2022)Chen, Zhang, Wang, and Gu]{chen2022}
Z.~Chen, L.~Zhang, X.~Wang, and P.~Gu.
\newblock Optimal design of flexible job shop scheduling under resource
  preemption based on deep reinforcement learning.
\newblock \emph{Complex System Modeling and Simulation}, 2\penalty0
  (2):\penalty0 174--185, 2022.

\bibitem[Creemers(2018)]{creemers2018}
S.~Creemers.
\newblock Maximizing the expected net present value of a project with
  phase-type distributed activity durations: An efficient globally optimal
  solution procedure.
\newblock \emph{European Journal of Operational Research}, 267\penalty0
  (1):\penalty0 16–22, 2018.

\bibitem[Creemers et~al.(2015)Creemers, De~Reyck, and Leus]{creemers2015}
S.~Creemers, B.~De~Reyck, and R.~Leus.
\newblock Project planning with alternative technologies in uncertain
  environments.
\newblock \emph{European Journal of Operational Research}, 242\penalty0
  (2):\penalty0 465–476, 2015.

\bibitem[Elmaghraby and Herroelen(1990)]{elmaghraby1990}
S.~E. Elmaghraby and W.~S. Herroelen.
\newblock The scheduling of activities to maximize the net present value of
  projects.
\newblock \emph{European Journal of Operational Research}, 49\penalty0
  (1):\penalty0 35--49, 1990.

\bibitem[Grinold(1972)]{grinold1972}
R.~C. Grinold.
\newblock The payment scheduling problem.
\newblock \emph{Naval Research Logistics Quarterly}, 19\penalty0 (1):\penalty0
  123--136, 1972.

\bibitem[Hermans and Leus(2018)]{hermans2018}
B.~Hermans and R.~Leus.
\newblock Scheduling markovian pert networks to maximize the net present value:
  New results.
\newblock \emph{Operations Research Letters}, 46\penalty0 (2):\penalty0
  240–244, 2018.

\bibitem[Herroelen et~al.(1997)Herroelen, Van~Dommelen, and
  Demeulemeester]{herroelen1997}
W.~S. Herroelen, P.~Van~Dommelen, and E.~L. Demeulemeester.
\newblock Project network models with discounted cash flows a guided tour
  through recent developments.
\newblock \emph{European Journal of Operational Research}, 100\penalty0
  (1):\penalty0 97--121, 1997.

\bibitem[Leyman and Vanhoucke(2016)]{leyman2016}
P.~Leyman and M.~Vanhoucke.
\newblock Payment models and net present value optimization for
  resource-constrained project scheduling.
\newblock \emph{Computers \&amp; Industrial Engineering}, 91:\penalty0
  139–153, 2016.

\bibitem[Leyman and Vanhoucke(2017)]{leyman2017}
P.~Leyman and M.~Vanhoucke.
\newblock Capital- and resource-constrained project scheduling with net present
  value optimization.
\newblock \emph{European Journal of Operational Research}, 256\penalty0
  (3):\penalty0 757–776, 2017.

\bibitem[Liang et~al.(2018)Liang, Cui, Wang, and Demeulemeester]{liang2018}
Y.~Liang, N.~Cui, T.~Wang, and E.~Demeulemeester.
\newblock Robust resource-constrained max-npv project scheduling with
  stochastic activity duration.
\newblock \emph{OR Spectrum}, 41\penalty0 (1):\penalty0 219–254, 2018.

\bibitem[Luo(2020)]{luo2020}
S.~Luo.
\newblock Dynamic scheduling for flexible job shop with new job insertions by
  deep reinforcement learning.
\newblock \emph{Applied Soft Computing}, 91:\penalty0 106208, 2020.

\bibitem[Mnih et~al.(2015)Mnih, Kavukcuoglu, Silver, Rusu, Veness, Bellemare,
  Graves, Riedmiller, Fidjeland, Ostrovski, Petersen, Beattie, Sadik,
  Antonoglou, King, Kumaran, Wierstra, Legg, and Hassabis]{mnih2015}
V.~Mnih, K.~Kavukcuoglu, D.~Silver, A.~A. Rusu, J.~Veness, M.~G. Bellemare,
  A.~Graves, M.~Riedmiller, A.~K. Fidjeland, G.~Ostrovski, S.~Petersen,
  C.~Beattie, A.~Sadik, I.~Antonoglou, H.~King, D.~Kumaran, D.~Wierstra,
  S.~Legg, and D.~Hassabis.
\newblock Human-level control through deep reinforcement learning.
\newblock \emph{Nature}, 518\penalty0 (7540):\penalty0 529–533, 2015.

\bibitem[Mohaghar et~al.(2016)Mohaghar, Khoshghalb, Rajabi, and
  Khoshghalb]{mohaghar2016}
A.~Mohaghar, A.~Khoshghalb, M.~Rajabi, and A.~Khoshghalb.
\newblock Optimal delays, safe floats, or release dates? applications of
  simulation optimization in stochastic project scheduling.
\newblock \emph{Procedia Economics and Finance}, 39:\penalty0 469–475, 2016.

\bibitem[Peymankar et~al.(2021)Peymankar, Davari, and Ranjbar]{peymankar2021}
M.~Peymankar, M.~Davari, and M.~Ranjbar.
\newblock Maximizing the expected net present value in a project with uncertain
  cash flows.
\newblock \emph{European Journal of Operational Research}, 294\penalty0
  (2):\penalty0 442–452, 2021.

\bibitem[Phuntsho and Gonsalves(2024)]{phuntsho2024}
T.~Phuntsho and T.~Gonsalves.
\newblock Maximizing net present value for resource constraint project
  scheduling problems with payments at event occurrences using approximate
  dynamic programming.
\newblock \emph{Algorithms}, 17\penalty0 (5):\penalty0 180, 2024.

\bibitem[Rezaei et~al.(2020)Rezaei, Najafi, and Ramezanian]{rezaei2020}
F.~Rezaei, A.~A. Najafi, and R.~Ramezanian.
\newblock Mean-conditional value at risk model for the stochastic project
  scheduling problem.
\newblock \emph{Computers \&amp; Industrial Engineering}, 142:\penalty0 106356,
  2020.

\bibitem[Rostami et~al.(2024)Rostami, Creemers, and Leus]{rostami2024}
S.~Rostami, S.~Creemers, and R.~Leus.
\newblock Maximizing the net present value of a project under uncertainty:
  Activity delays and dynamic policies.
\newblock \emph{European Journal of Operational Research}, 317\penalty0
  (1):\penalty0 16--24, 2024.

\bibitem[Russell(1970)]{russell1970}
A.~Russell.
\newblock Cash flows in networks.
\newblock \emph{Management Science}, 16\penalty0 (5):\penalty0 357--373, 1970.

\bibitem[Sobel et~al.(2009)Sobel, Szmerekovsky, and Tilson]{sobel2009}
M.~J. Sobel, J.~G. Szmerekovsky, and V.~Tilson.
\newblock Scheduling projects with stochastic activity duration to maximize
  expected net present value.
\newblock \emph{European Journal of Operational Research}, 198\penalty0
  (3):\penalty0 697–705, 2009.

\bibitem[Waligóra(2008)]{waligra2008}
G.~Waligóra.
\newblock Discrete–continuous project scheduling with discounted cash
  flows—a tabu search approach.
\newblock \emph{Computers \&amp; Operations Research}, 35\penalty0
  (7):\penalty0 2141–2153, 2008.

\bibitem[Wang et~al.(2024)Wang, Lu, Qian, Hu, and Liu]{wang2024}
X.~Wang, S.~Lu, X.~Qian, C.~Hu, and X.~Liu.
\newblock Dynamic scheduling of decentralized high-end equipment r\&d projects
  via deep reinforcement learning.
\newblock \emph{Computers \& Industrial Engineering}, 190:\penalty0 110018,
  2024.

\bibitem[Wiesemann and Kuhn(2015)]{wiesemann2015}
W.~Wiesemann and D.~Kuhn.
\newblock The stochastic time-constrained net present value problem.
\newblock \emph{Handbook on project management and scheduling vol. 2}, pages
  753--780, 2015.

\bibitem[Wiesemann et~al.(2010)Wiesemann, Kuhn, and Rustem]{wiesemann2010}
W.~Wiesemann, D.~Kuhn, and B.~Rustem.
\newblock Maximizing the net present value of a project under uncertainty.
\newblock \emph{European Journal of Operational Research}, 202\penalty0
  (2):\penalty0 356–367, 2010.

\bibitem[Yao et~al.(2024)Yao, Tam, Wang, Le, and Butera]{yao2024}
Y.~Yao, V.~W. Tam, J.~Wang, K.~N. Le, and A.~Butera.
\newblock Automated construction scheduling using deep reinforcement learning
  with valid action sampling.
\newblock \emph{Automation in Construction}, 166:\penalty0 105622, 2024.

\bibitem[Zhang et~al.(2020)Zhang, Song, Cao, Zhang, Tan, and Chi]{zhang2020}
C.~Zhang, W.~Song, Z.~Cao, J.~Zhang, P.~S. Tan, and X.~Chi.
\newblock Learning to dispatch for job shop scheduling via deep reinforcement
  learning.
\newblock \emph{Advances in neural information processing systems},
  33:\penalty0 1621--1632, 2020.

\bibitem[Zheng et~al.(2018)Zheng, He, Wang, and Jia]{zheng2018}
W.~Zheng, Z.~He, N.~Wang, and T.~Jia.
\newblock Proactive and reactive resource-constrained max-npv project
  scheduling with random activity duration.
\newblock \emph{Journal of the operational research society}, 69\penalty0
  (1):\penalty0 115--126, 2018.

\end{thebibliography}

\label{sec:reference}
\end{document}